\RequirePackage{fix-cm}
\documentclass[smallextended]{svjour3}       
\smartqed  
\usepackage{graphicx}
\usepackage{subfig}
\usepackage{algorithm}
\usepackage{algorithmic}
\usepackage{url}
\usepackage{array}
\usepackage{rotfloat}
\usepackage{multirow}
\usepackage{rotating}
\usepackage{amsmath}
\usepackage{graphics}
\usepackage{caption}
\usepackage{booktabs}
\usepackage{amssymb}
\usepackage{xcolor}
\begin{document}

\title{Development of an Automatic 3D Human Head Scanning-Printing System
}



\author{Longyu Zhang \and Bote Han \and \\Haiwei Dong \and Abdulmotaleb El Saddik}

\institute{ \at Multimedia Computing Research Laboratory (MCRLab), School of Electrical Engineering and Computer Science, University of Ottawa, Ottawa, Ontario, Canada\\
Tel.: +1-613-562-5800 ext. 2148\\	
\email{\{lzhan121, bhan097, hdong, elsaddik\}@uottawa.ca}   }

\date{Received: date / Accepted: date}

\maketitle

\begin{abstract}
Three-dimensional (3D) technologies have been developing rapidly recent years, and have influenced industrial, medical, cultural, and many other fields. In this paper, we introduce an automatic 3D human head scanning-printing system, which provides a complete pipeline to scan, reconstruct, select, and finally print out physical 3D human heads. To enhance the accuracy of our system, we developed a consumer-grade composite sensor (including a gyroscope, an accelerometer, a digital compass, and a Kinect v2 depth sensor) as our sensing device. This sensing device is then mounted on a robot, which automatically rotates around the human subject with approximate 1-meter radius, to capture the full-view information. The data streams are further processed and fused into a 3D model of the subject using a tablet located on the robot. In addition, an automatic selection method, based on our specific system configurations, is proposed to select the head portion. 
We evaluated the accuracy of the proposed system by comparing our generated 3D head models, from both standard human head model and real human subjects, with the ones reconstructed from FastSCAN and Cyberware commercial laser scanning systems through computing and visualizing Hausdorff distances. Computational cost is also provided to further assess our proposed system.
\end{abstract}

 \section{Introduction}
\label{intro}

The rise of 3D digitization technologies has enabled researchers and engineers from a wide range of fields \cite{Chen_2016,Zhang_2015} to use accurate human models for many practical applications, such as anthropological studies, digital avatar animation and ergonomic product design. In anthropological studies, researchers have been investigating the relationship between facial shape variations and neurological and psychiatric disorders. For example, Hennesy et al. used 3D head models acquired from laser scanners to identify schizophrenia from facial dysmorphic features \cite{Hennessy-2002}. A fast algorithm for 3D face reconstruction with uncalibrated photometric stereo technology was also proposed by Qi et al. \cite{mtap3}.  

Human avatar animation has also become popular with the development of 3D graphics and gaming. Lee and Magnenat-Thalman introduced a method to reconstruct 3D facial models for animation from two orthogonal images (frontal and profile view) or from range data \cite{Lee-2000}. Additionally, Kan and Ferko adopted this same principle to build an automatic system where they use the facial feature matching of two images and a parametrized head model to create 3D head models as avatars in 3D games \cite{Kan-2010}.

An important part of 3D human model is head model, which can be used to establish standards for the design of products that fit onto the face or head, such as respiratory masks, glasses, helmets or other head-mounted devices \cite{Friess-2003}. An interesting initiative was the Size-China project \cite{Ball-Innovation2008,Ball-CAD2011}. To find the proper fit for Asians, who have different head shapes compared with Westerners in facial-head products such as helmets, face masks, and caps, and to derive standards with anthropometric database, Ball et al. created an Asian anthropometric database built from 3D scans of 2000 Asian people using a stationary head and face color 3D scanner by Cyberware\footnote{Head \& Face Color 3D Scanner (Model PX) \url{http://www.cyberware.com/}}, from which several standard Asian head and face models were created. These types of surveys are essential for global product design, as the anthropometric properties of body parts vary from culture to culture.

\begin{figure*}[t]
	\centering
	\includegraphics[width=.94\linewidth]{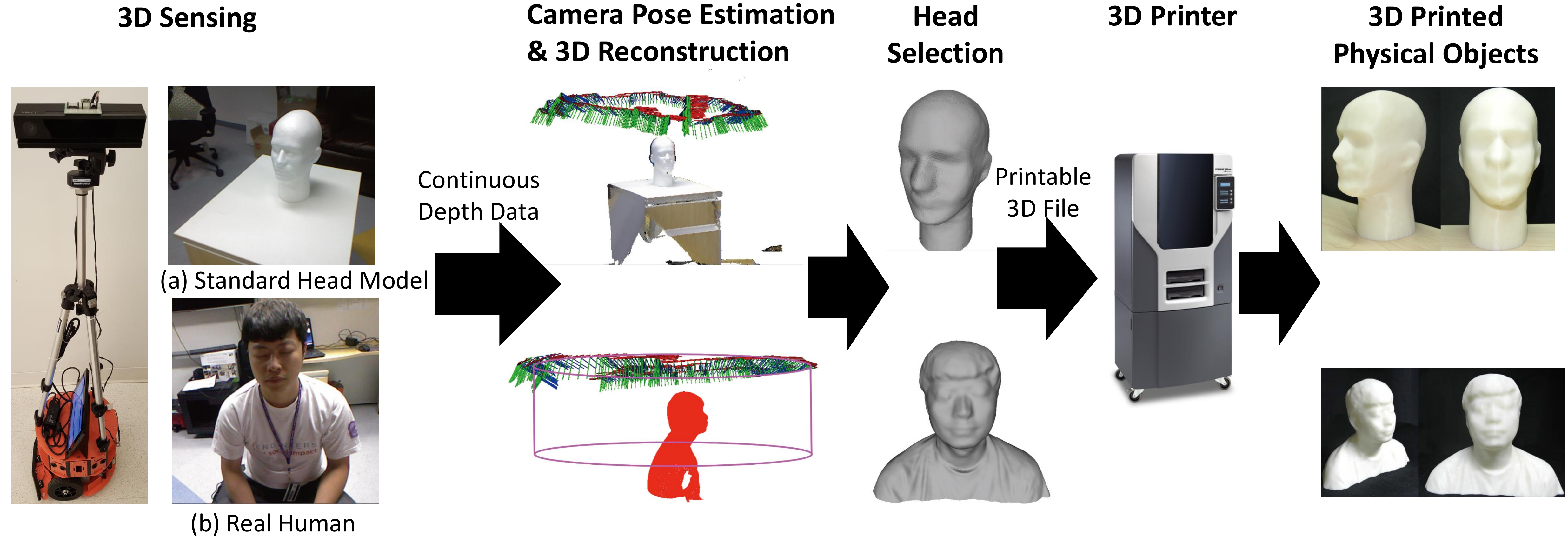}
	\caption{System Architecture.}
	\label{fig:flowchart}
\end{figure*}

Most previously mentioned applications require many manual steps, either to build the model, select head model, or clean it up. Besides, some of them rely on expensive scanning devices, such as 3D laser scanner. Therefore, in this paper, we introduce an automatic 3D human head scanning-printing system, which provides a complete pipeline to scan, reconstruct, select, and generate 3D head model to 3D printer. Our system architecture is shown as Fig. \ref{fig:flowchart}.
We utilize our developed composite sensing device, carried by a robot which automatically rotates around the subject with approximate 1-meter radius, to capture full-view information, and then fuses them to reconstruct 3D model of the subject with a tablet located on the robot. An automatic selection method based on sensor poses estimation and head matrix transformation is further presented to select the 3D human head. Both physical standard human head model and real human subjects work well with our system. Furthermore, the proposed system is mainly based on consumer-grade sensors, which facilitates the acquisition of 3D human heads to common users, and enables projects such as Size-China to reach larger global dimensions. Furthermore, the output of our proposed system can be used to print out prototypes of human heads with the help of 3D printing technologies.

In this paper, our main contributions are as follows:
	\begin{itemize}
		\item Through combining a gyroscope, an accelerometer, a digital compass, and a Kinect v2 depth sensor into one consumer-grade composite sensing device, and mounting it on a mobile robot, an automatic scanning device is developed to rotate around the human subject with approximate 1-meter radius, and capture full-view information. Global locations and poses of Kinect v2 are estimated real-time from both depth images matching results and motion data from the gyroscope, digital compass and accelerometer. Utilizing hardware and software simultaneously enables to obtain accurate depth sensor's location and orientation, and further improves the 3D reconstruction accuracy.
		     
		\item The 3D reconstruction results of our proposed low-cost system are compared with the results from commercial laser scanning systems: FastSCAN and Cyberware. Through computing and visualizing Hausdorff distance comparison, our cost-effective system is proved to achieve compelling results.
		
		
		\item A complete pipeline for an automatic 3D human head scanning-printing system is introduced, which can scan, reconstruct, select, and finally print out a physical 3D human heads.
	\end{itemize}

This paper is organized as follows. Section 2 presents an overview of the research areas related to 3D sensing, 3D reconstruction, and 3D printing. In Section 3, we illustrate the proposed system and give detailed descriptions of each component. Section 4 addresses the implementation details and discusses the experimental results. Finally, we provide our conclusions and suggestions for future work in Section 5.

\section{Related Works}
\label{sec:Background}

The reconstruction of human heads has been widely researched for the past ten years, with the aim of producing various applications described in the previous section. Several approaches, such as photogrammetry \cite{Photo1} and Fourier transform profilometry \cite{SuChen-2001}, have acquired compelling results recent years. In our proposed system, we developed a consumer-grade composite sensing device to acquire data streams, realized our 3D reconstruction process through modifying KinectFusion algorithm \cite{015KinectFusion1}, presented our head selection method, and physically printed the generated 3D head. In order to further understand our reasoning for the selection of device and methods, we present the following overview regarding 3D sensing devices, 3D reconstruction methods, and 3D printers.

\subsection{3D Sensing Devices}

The goal of 3D sensing device is to generate 3D representations of the world from the viewpoint of a sensor \cite{Park_2012}. These 3D representations are generally in the form of 3D point clouds or polygon meshes \cite{mtap4}. Each point $p$ of a 3D point cloud has $(x,y,z)$ coordinates relative to the fixed coordinate system of the origin of the sensor \cite{Zhang_2016,Zhang_2016_1}. Depending on the sensing device the point $p$ can, additionally contain color information, such as $(r,g,b)$ values. Commonly used sensors to reconstruct 3D models include \textit{passive-image sensors}, \textit{laser sensors}, \textit{structured-light sensors}, and \textit{time-of-flight sensors}:

\textit{Passive-Image-Based  Sensors} usually use a set of traditional 2D cameras, which do not emit any kind of radiation themselves and are only capable to capture 2D images without depth information \cite{004Semi1}. However, researchers can still reconstruct 3D models with photogrammetry methods, which are based on simple or multi-triangulation principle between homologous optical rays departing from the object and reaching the image sensor. For example, Remondino et al. introduced a dense reconstruction method by applying multi-image high density image matching \cite{Photo1}; and De Souza et al. used a photogrammetric method to detect newborn infants' head surface shape \cite{Photo2}.        

\textit{Triangulation-Based Laser Sensors} usually emit a laser on the target and employ a camera to detect the location of the laser dot, and then, based on the distance that the laser strikes a surface, the laser dot will appear at different places within the camera's field of view. This method is called triangulation because the camera, the laser dot, and the laser emitter form into a triangle \cite{161Laser}. These kinds of sensors are generally able to acquire data with high-quality to build precise 3D models, but are usually more expensive compared with other types of sensors and they require expert knowledge to operate. Examples of this kind of sensors include Cyberware Whole Body Color 3D Scanner, NextEngine desktop 3D scanner, and Creaform's handheld HandyScan scanner. Moreover, users always need
to sit still during the capturing process, which is difficult in
certain situations, such as sensing 3D models for infants \cite{072Space}.

\textit{Structured-Light Sensors} project patterns consisting of many stripes at once, or of arbitrary finges, and allow the acquisition of a multitude of samples simultaneously. Microsoft's Kinect v1 for Xbox 360 and for Windows, released in 2010 and 2012 separately, are adopting this technology. The availability of KinectFusion \cite{015KinectFusion1}, a real-time 3D reconstruction and interaction algorithm, further exploded the utilization of Kinect v1 as a 3D sensor. Many 3D reconstruction applications, such as KScan3D\footnote{KScan3D. \url{http://www.kscan3d.com/}} and ReconstructMe\footnote{ReconstructMe. \url{http://reconstructme.net/}}, have been developed with Kinect v1 and KinectFusion.

\textit{Time-of-Flight Sensors} are different from structured-light
sensors in working principles. ToF sensors actively measure the distance of a surface by recording the round-trip time
of the emitted infrared light, and commercially available ToF sensors commonly employ homodyning methods and operate within continuous mode \cite{054ToF}. ToF sensors occupy several capabilities, such as capturing depth images at
video rate under low-light levels, being color and texture invariance, and resolving silhouette ambiguities
in pose \cite{013Track}. Among ToF companies, MESA Imaging produced Swiss Ranger SR4k family\footnote{Swiss Ranger. http://www.mesa-imaging.ch/swissranger4500.php}, and Microsoft created Kinect v2.

Among aforementioned sensors, Microsoft Kinect v1 and v2 sensors have dramatically improved the 3D reconstruction research area. A comprehensive review of Kinect-based reconstruction algorithms and applications aiming addressing traditional challenges in this field has been conducted by Han et al. \cite{re3}. They outlined main research contributions in sparse feature matching and dense point matching methods, and presented that advanced machine learning techniques may be integrated to further improve the results. In our work, we opted to use the Microsoft Kinect v2 RGB-D sensor as our depth device for several reasons: Kinect v2 sensor has compact size, low consumer price, the capability to capture color and depth data at video rate, easy availability on the market, acceptable accuracy, and they operate safely for both the users and the scanned subjects  \cite{yang2015evaluating}. However, Kinect v2 suffers mobility limitations because of its USB cable and power cable. Thus, we utilized a tablet and a portable battery to overcome these problems.

\subsection{3D Model Reconstruction}

In this subsection, we address the challenge of reconstructing 3D models using multiple views or point clouds obtained by the 3D sensing device. These multiple views can be obtained by moving the 3D sensing device around the object until a full set of $360^\circ$ views of the object is obtained. Additionally, the object can be spun $360^\circ$ around its axis with the 3D sensing device fixed on a certain viewing point. The range images must overlap with the previous ones. Full 3D models of the objects can be reconstructed by registering these multiple views \cite{Merchan_2012,MTAP1}. Registration is the estimation of the rigid motion (translations and rotations) of a set of points with respect to another set of points. The rigid motion is estimated using the surface of the object that is common between successive "views". These can be image pixels or 3D points. This rigid motion is then estimated by using \emph{coarse} or \emph{fine} registration methods, as well as a combination of both. Coarse registration methods are RANSAC-based algorithms that use sparse feature matching, first introduced by Chen et al. \cite{Chen:1998} and Feldmar \cite{Feldmar:1994}. These generally provide an initial guess for fine registration algorithms, relying on minimizing point-to-point, point-to-plane, or plane-to-plane correspondences. Genetic Algorithms and Iterative Closest Point (ICP) are widely used to solve those problems \cite{Besl92,MTAP2,Zhengyou94}. In some cases, a coarse registration is not necessary, since the point clouds are already very close to each other or semi-aligned, a fine registration can then be further implemented.

For the problem of registering multiple point clouds, many possible approaches have been examined.  An offline multiple-view registration method has been introduced by Pulli \cite{Pulli99}. This method computes pair-wise registrations as an initial step and uses their alignments as constraints for a global optimization step. This global optimization step registers the complete set of point clouds simultaneously and diffuses the pair-wise registration errors. A similar approach was also presented by Nishino et al. \cite{Nishino02}. Chen et al. \cite{Chen91} developed a metaview approach to register and merge views incrementally. Masuda introduced a method to bring pre-registered point clouds into fine alignment using the signed distance functions \cite{Masuda02}. A simple pair-wise incremental registration would suffice to obtain a full model if the views contain no alignment errors. This becomes a challenging task when dealing with noisy datasets. Some approaches use an additional offline optimization step to compensate for the alignment errors for the set of rigid transformations \cite{Weise09}.

All of these previously mentioned algorithms target raw or filtered data from the 3D sensing device (i.e., 3D points), which lack a resulting 3D model that has a tight surface representation of the object. Thus, in order to convert these 3D reconstructions to 3D CAD models, several post-processing steps need to be applied. Initially, the set of points need to be transformed into a water-tight (no hole) polygon mesh, this can be done by meshing algorithms. Some popular meshing algorithms include:  greedy triangulation \cite{Dickerson94fastgreedy}, marching cubes \cite{Lorensen87marchingcubes} and poisson reconstruction \cite{Kazhdan2006}.

Recently, Newcombe et al. introduced their novel reconstruction system, \emph{KinectFusion}, which fuses dense depth data streamed from a Kinect into a single global implicit surface model in real-time \cite{KinectFusion,015KinectFusion1}. They use a volumetric representation, called the truncated signed distance function (TSDF), and combine it with a fast ICP (Iterative Closest Point). The TSDF representation is suitable for generating 3D CAD models. In other words, in this approach the surface is extracted beforehand (with the TSDF representation), then the classical ICP-based registration approach is performed to generate the full reconstruction. A commercial application released soon after Kinectfusion is previously mentioned ReconstructMe. This software is based on the same principal of incrementally aligning a TSDF from Kinect data on a dedicated GPU.

Several researches have been conducted to deal with certain circumstances. For example, Shum et al. \cite{re2} proposed a method to solve the occlusion problems. Their method improved the human subject recognition accuracy, and further optimized the posture reconstruction results. To overcome the restrictions that the subject must stay still during the entire scanning process, Barmpoutis \cite{re1} presented a method of reconstructing moving human subjects with RGB-D frames real-time. He proposed a method to estimate the positive-definite tensor-splines, and obtained compelling results.

Since we target at an accurate reconstruction system, we require the user sit still during the scanning process. Original KinectFusion algorithm is mainly based on modified ICP methods, which may introduce accumulative errors. Thus, we adopted a gyroscope, a digital compass,  and an accelerometer to provide additional tracking information of Kinect v2. Then the final sensor locations and poses are calculated from both ICP matching results and hardware tracking results with proper weighting values. Furthermore, the main head portion of the reconstructed mesh can be extracted with our proposed automatic selection method, based on our specific system configurations.  
	

\subsection{3D Printers}

3D printing is an additive technology in which 3D objects are created using layering techniques of different materials, such as plastic, metal, etc. The first 3D printing technology developed in the 1980's was stereolithography (SLA) \cite{SLA}. This technique uses an ultraviolet (UV) curable polymer resin and an UV laser to build each layer one by one. Since then numerous 3D printing technologies have been introduced \cite{3DSurvey}. For example, the polyjet technology, which works like an inkjet document printer instead of jetting drops of ink, jets layers of liquid photopolymer and cures them with a UV light. Another 3D printing technology is fused deposition modeling (FDM), based on material extrusion, a thermoplastic material is heated up into semi-liquid state and extruded from a computer-controlled print head. This FDM technology has become specifically popular for commercial 3D printers.

\section{The Proposed System}
\label{sec:SystemArch}
\label{sec:Implementation}


In this section, we illustrate the details of our proposed system. The system architecture is shown in Fig. \ref{fig:flowchart}. The developed consumer-grade composite sensor (including Kinect v2 sensor, gyroscope, digital compass, and accelerometer), is mounted on a robot which automatically rotates around the subject with approximate 1-meter radius. This sensing device captures depth images of a human subject in a $360^\circ$ fashion. The depth images are then processed and fused into a globally implicit surface model based on both KinectFusion algorithm and the additional sensor locations and poses provided by the sensing device with proper weighting values. Computation of data steams are processed by a tablet placed on the robot for enhanced mobility. Our system then creates a virtual plane to select the main head portion. Finally, the reconstructed head model is printed out by a 3D printer.


\subsection{Proposed Scanning Device}

	
		\begin{figure}[t]
			\centering
			\includegraphics[width=.8\linewidth]{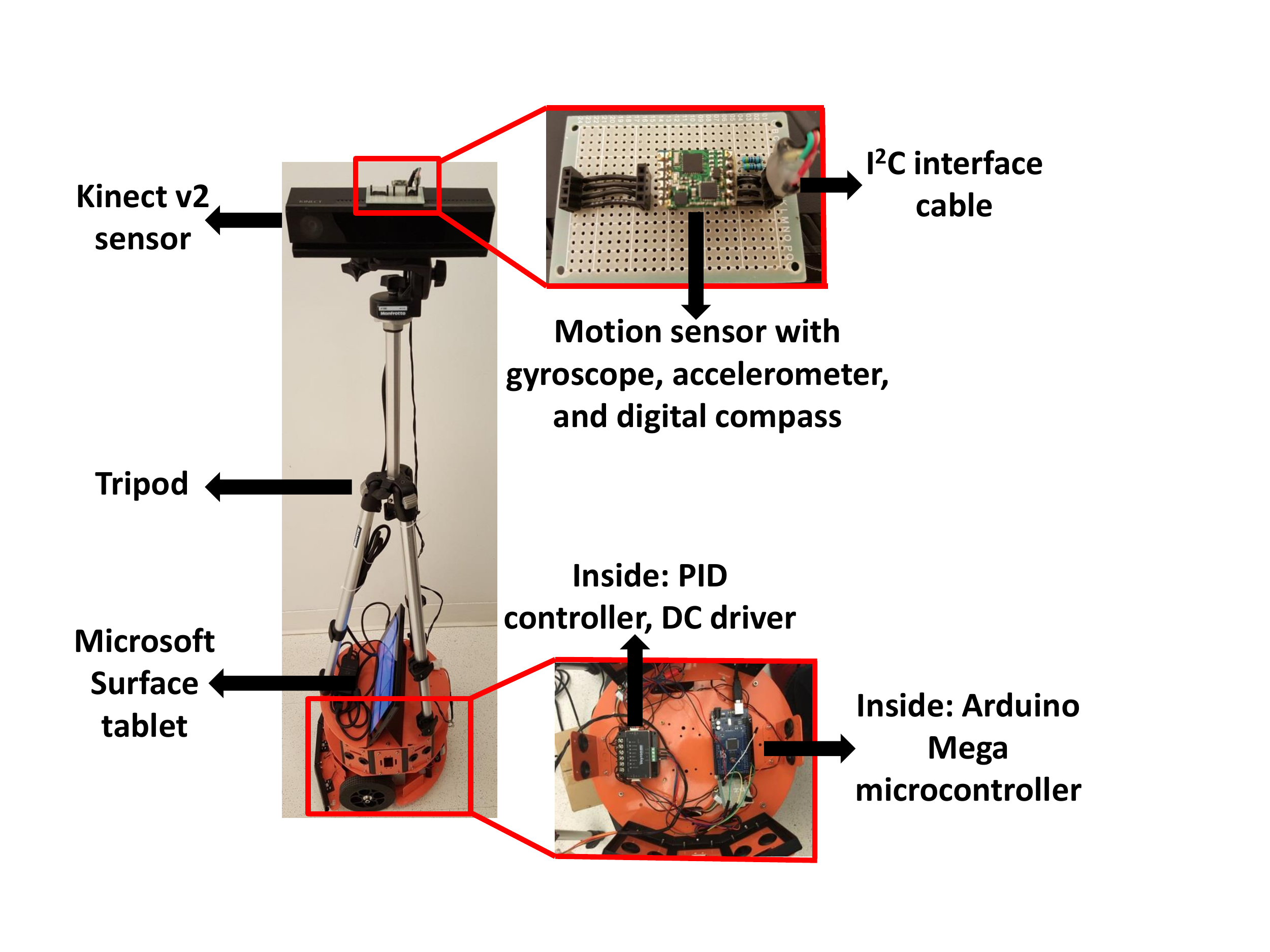}
			\caption{Developed scanning device with a composite sensor, tripod, Microsoft Surface tablet, and robot.}
			\label{fig:hs}
		\end{figure}

	Our proposed scanning device is show in Fig. \ref{fig:hs}, which contains a composite sensor, a tripod, a Microsoft Surface tablet, and a robot. As mentioned before, we choose Microsoft Kinect v2 sensor as our depth sensor. With time-of-flight range detection technology, it can observe depth images of the subject with detailed distance information.
	 Because Kinect v2 requires extra power source, we cut off its power adaptor cable and added an extra portable battery to increase the mobility.
	 To reduce the accumulative errors generated from fusing depth images, we attached a gyroscope, an accelerometer, and a digital compass on top of Kinect v2 to develop a composite sensor with additional global locations and poses information. Microelectromechanical gyroscope is able to monitor the pose rotations of Kinect v2, but it is sensitive to environments and may cause drifting problems. Therefore, we applied Kalman filter algorithm \cite{Kalman}, which uses a series of measurements to provide the optimal estimation results, to reduce the noise, and fused it with the data from a digital compass to achieve an angle precision as high as 0.01 degree without drifting problem. An accelerometer was also adopted to measure the moving accelerations, and was combined with previously obtained angle information to calculate the depth sensor locations and poses.
	
	In order to realize automatic scanning of the human subject, we mounted our composite sensor on a tripod, and fixed them with a robot, which is programmed to rotate around the subject with approximate 1-meter radius. The robot we adopted is DFRobot's HCR Mobile Robot\footnote{HCR Mobile Robot: http://www.dfrobot.com}, a two-wheel drive platform. We used Arduino Mega microcontroller to control the robot movements and collect motion data from both the composite sensor and the robot itself. The robot comes with two 12V direct-current (DC) geared motors. Its reduction ratio is 51:1 with the encoder resolution of 663 pulses per round. Since DC motors may generate inconsistent motion, we utilized a proportional-integral-derivative (PID) controller to improve the precision. After several testing, we set the left-wheel speed as 13 rounds per minute and the right-wheel speed as 16 rounds per minute. The microcontroller communicates with the motor driver through serial communications, while receives data from the composite sensor by I$^2$C interface.
	
	For mobility and proper payload, we used a tablet as our computational resource, instead of a laptop. In our experiment, we adopted Microsoft Surface tablet to fulfill Kinect v2's high demanding for graphic processing units and USB connection. It can also be easily placed on the robot as shown in Fig. \ref{fig:hs}. Since both Kinect v2 and the robot microcontroller interact with the tablet through USB serial connections, we use a USB 3.0 four-port hub from Unitek to extend the tablet's single USB port into multiple ones. Thus, the tablet can successfully receive the data streams and process them.

\subsection{3D Model Reconstruction}

KinectFusion is based on incrementally fusing consecutive frames of depth data into a 3D volumetric representation of an implicit surface. This is a representation of the truncated signed distance function (TSDF) \cite{SDF}. The TSDF is basically a 3D point cloud stored in GPU memory using a 3D voxelized grid. The global TSDF is updated every time a new depth image frame is acquired and the current camera pose with respect to the global model is estimated. Initially, the depth image from the Kinect v2 sensor is smoothed out using a bilateral filter \cite{eccv-12-qingxiong-yang}, which up-samples the raw data and fills the depth discontinuities. Then the camera pose of the current depth image frame is estimated with respect to the global model by applying a fast Iterative-Closest-Point (ICP) algorithm between the currently filtered depth image and a predicted surface model of the global TSDF extracted by ray casting. Once the camera pose is estimated, the current depth image is transformed into the coordinate system of the global TSDF and updated. In following parts, we illustrate the details of our method.


	\subsubsection{Camera Pose Estimation}

	The principal of the ICP algorithm is to find a data association between the subset of the source points ($P_{s}$) and the subset of the target points ($P_{t}$) \cite{Besl92,Chen91}. Let's define a homogenous transformation
	$T$ of a point in $P_{s}$ (denoted as $p_{s}$)  with respect to a point in $P_{t}$ (denoted as $p_{t}$) as
	
	\begin{equation}
	p_{t}=T\left(p_{s}\right)=\left[\begin{array}{cc}
	R & t\\
	0 & 1
	\end{array}\right]p_{s}
	\end{equation}
	where $R$ is a rotational matrix and $t$ is a translational vector.
	Thus, ICP can be formulated as
	
	
	\begin{equation}
	T^{*} = \underset{T}{\mathrm{argmin}}\sum_{p_{s}\in P_{s}}\left(T\left(p_{s}\right)-p_{t}\right)^{2} = \underset{T}{\mathrm{argmin}}\sum\left\Vert T\left(P_{s}\right)-P_{t}\right\Vert^2
	\end{equation}
	
	A special variant of the ICP-algorithm, the point-to-plane ICP  \cite{Zhengyou94}, is used. It minimizes the error along the surface normal of the target points $n_t$, as in the following equation:
	\begin{equation}
	T^{*}  =  \underset{T}{\mathrm{argmin}}\sum_{p_{s}\in P_{s}} \lVert n_t \cdot (T(p_{s})-p_{t})\rVert_{2}\\
	\end{equation}
	where $n_t \cdot (T(p_{s})-p_{t})$ is the projection of $(T(p_{s})-p_{t})$ onto the sub-space spanned by the surface normal ($n_t$). After computing transformation $T$, the new depth image is transformed into the global coordinate system.

	
	Since ICP may introduce accumulative errors, we utilize aforementioned gyroscope, digital compass, and accelerometer to track additional information, i.e., Kinect v2's location and orientation. The redundant information from hardware are then fused with the software calculation results to improve the system's accuracy and robustness. Because ICP itself can realize acceptable matching results mostly \cite{ICP3}, we give it a large weighting value (0.8), while set the weighting value from hardware smaller (0.2). Combining both hardware and software information reduces the accumulative errors, while remaining the advantages of original KinectFusion algorithm.

	\subsubsection{Global TSDF Updating}
	\label{sssec:global}
	
	The global model is represented in a voxelized 3D grid and integrated using a simple weighted running average. For each voxel, we have a value of signed distance for a specific voxel point $x$ as $d_{1}\left(x\right)$, $d_{2}\left(x\right)$, $\cdots$, $d_{n}\left(x\right)$ from $n$ depth images ($d_i$) in a short time interval. To fuse them, we define $n$ weights $w_{1}\left(x\right)$, $w_{2}\left(x\right)$, $\cdots$, $w_{n}\left(x\right)$. Thus, the weight corresponding point matching can be written in the form
	\begin{equation}
	w_{n}^{*}=\underset{k}{\mathrm{arg}}\sum_{k=1}^{n-1}\left\Vert W_{k}D_{k}-D_{n}\right\Vert _{2}
	\end{equation}
	where
	\begin{eqnarray}
	D_{k+1} & = & \frac{W_{k}D_{k}+w_{k+1}d_{k+1}}{W_{k}+w_{k+1}}\\
	W_{k+1} & = & W_{k}+w_{k+1}
	\end{eqnarray}
	$D_{k+1}$ is the cumulative TSDF and $W_{k+1}$ is the weight functions after the integration of the current depth image frame. Furthermore, by truncating the update weights to a certain value $W_{\alpha}$ a moving average reconstruction is obtained.
	
	\subsubsection{Meshing with Marching Cubes}
	\label{sssec:marchingcubes}
	
\begin{figure}[t]
		\centering
		\subfloat[3D point cloud.] 
		{\includegraphics[height=3cm]{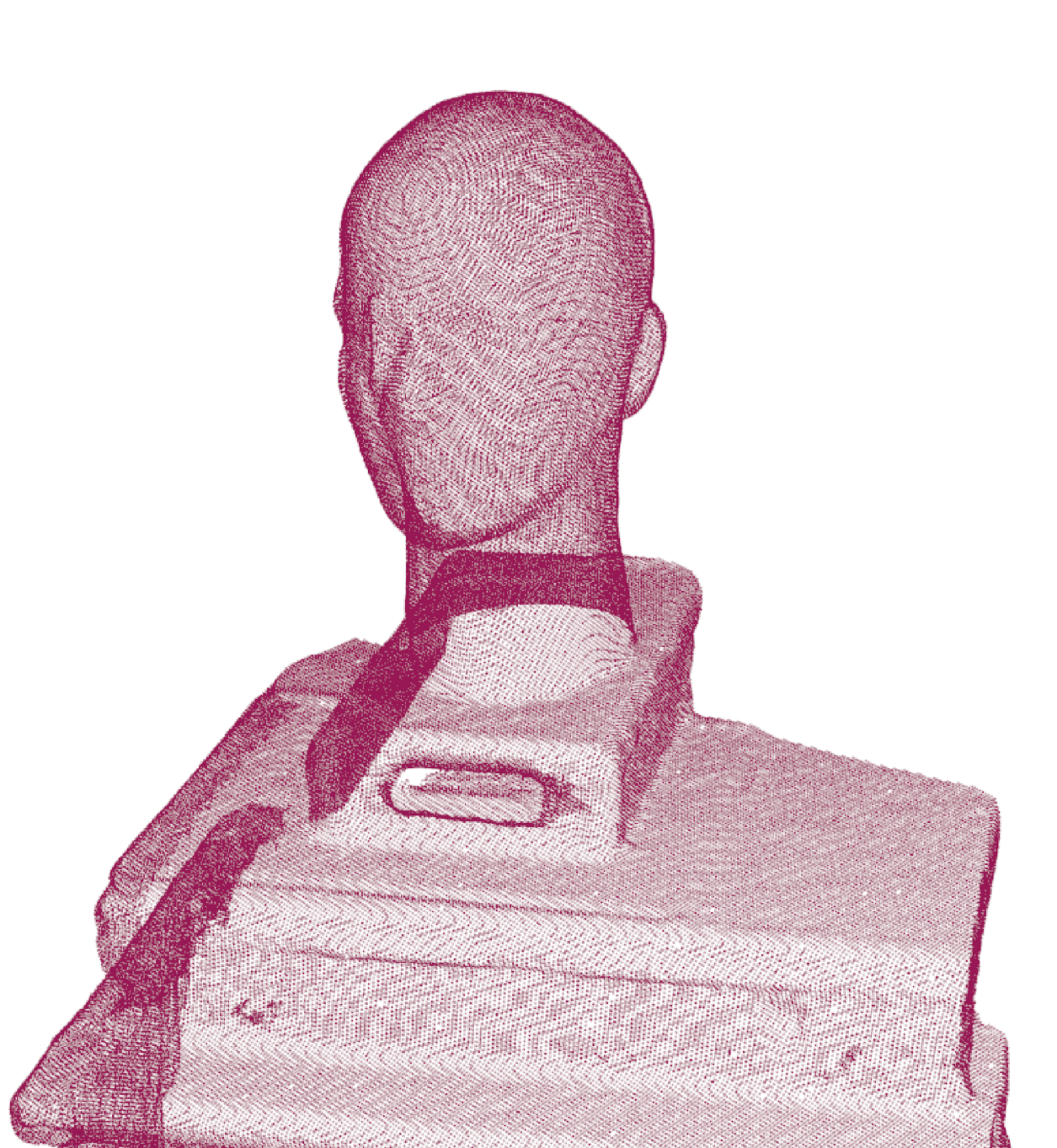}} \qquad %
		\subfloat[Mesh.]
		{\includegraphics[height=3cm]{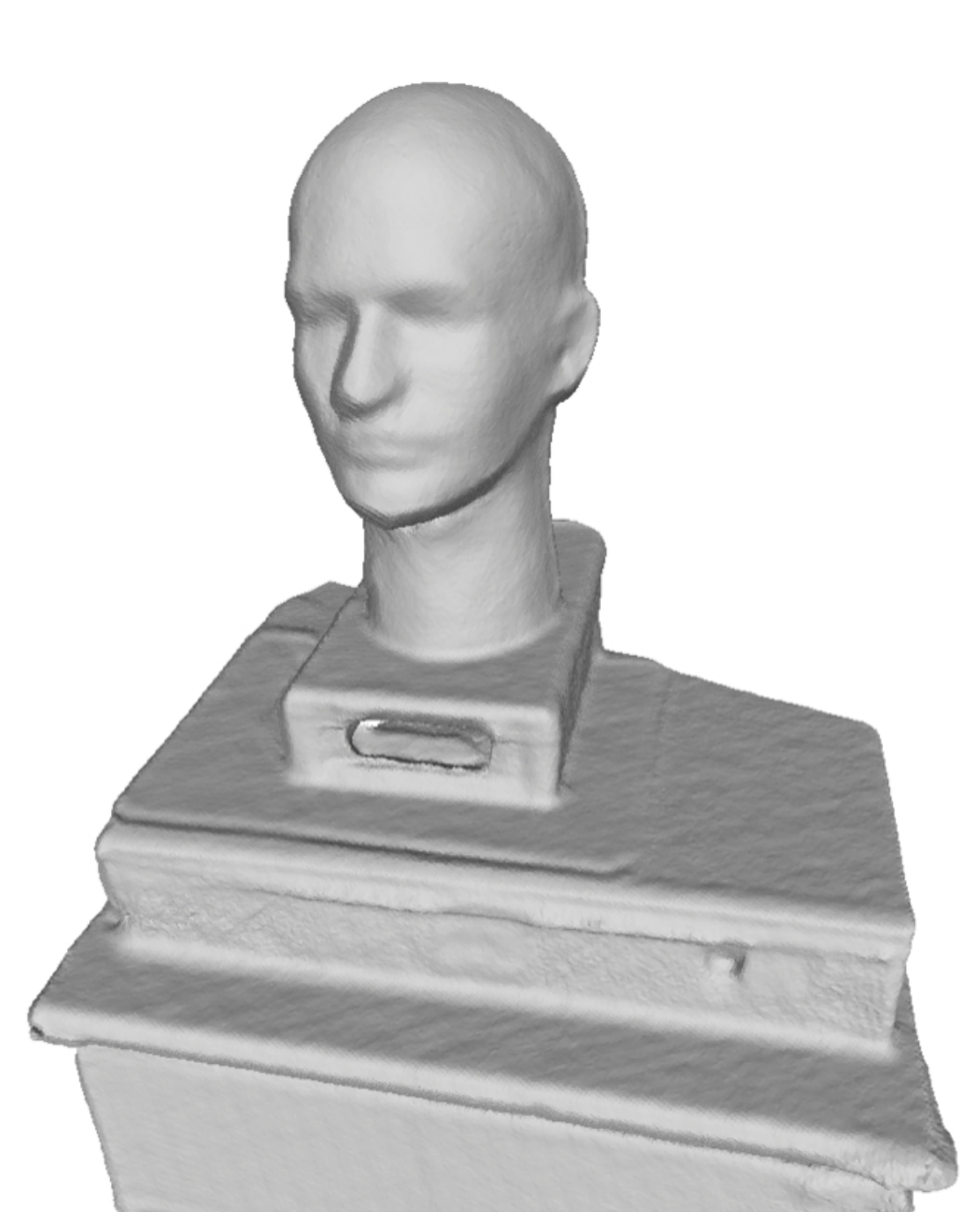}}
		\caption{3D reconstructed standard human head model. (a) point cloud format (b) polygon mesh.}
		\label{fig:cloud2mesh}
	\end{figure}

	The final global TSDF can be converted to a point cloud or polygon mesh representation. The polygon mesh is extracted by applying the marching cubes algorithm to the voxelized grid representation of the 3D reconstruction \cite{Lorensen87marchingcubes}. The marching cubes algorithm extracts a polygon mesh by subdividing the points cloud or set of 3D points into small cubes (voxels) and marching through each of these cubes to set polygons that represent the isosurface of the points lying within the cube. This results in a smooth surface that approximates the isosurface of the voxelized grid representation. In Fig. \ref{fig:cloud2mesh}, a successful reconstruction of a human head can be seen. We walked around the object closing a $360^{\circ}$ loop. The resulting polygon mesh is to be used as CAD model to virtualize the scanned object for 3D printing.
	
	However, as shown in Fig. \ref{fig:cloud2mesh}, this final 3D reconstructed model includes portions of the table or the environment that we are not willing to print or visualize. This holds for scans of people as well. When we create scans of humans and want to print a 3D head, we need to manually trim the 3D models. This is an obstacle for trying to create an automatic 3D sensing to printing system. Thus, in the next subsection we present our proposed approach for automatic 3D model post-processing which generates a \emph{ready-to-print} polygon mesh by applying selection method to the 3D point cloud of the reconstructed models.

\subsection{3D Head Selection}

Our head selection method is based on two specific assumptions of our designed system. The first one is that the scanned standard human head model or real human head is standing/sitting/lying on a plane. The second assumption is that the 3D sensing device is approximately close a loop around this object, namely, the robot must rotate around the scanned subject in a $360^{\circ}$ fashion.


\begin{figure}[htbp]
	\centering
	\subfloat[Full reconstructed model with sensor poses] 
	{\includegraphics[height=3cm]{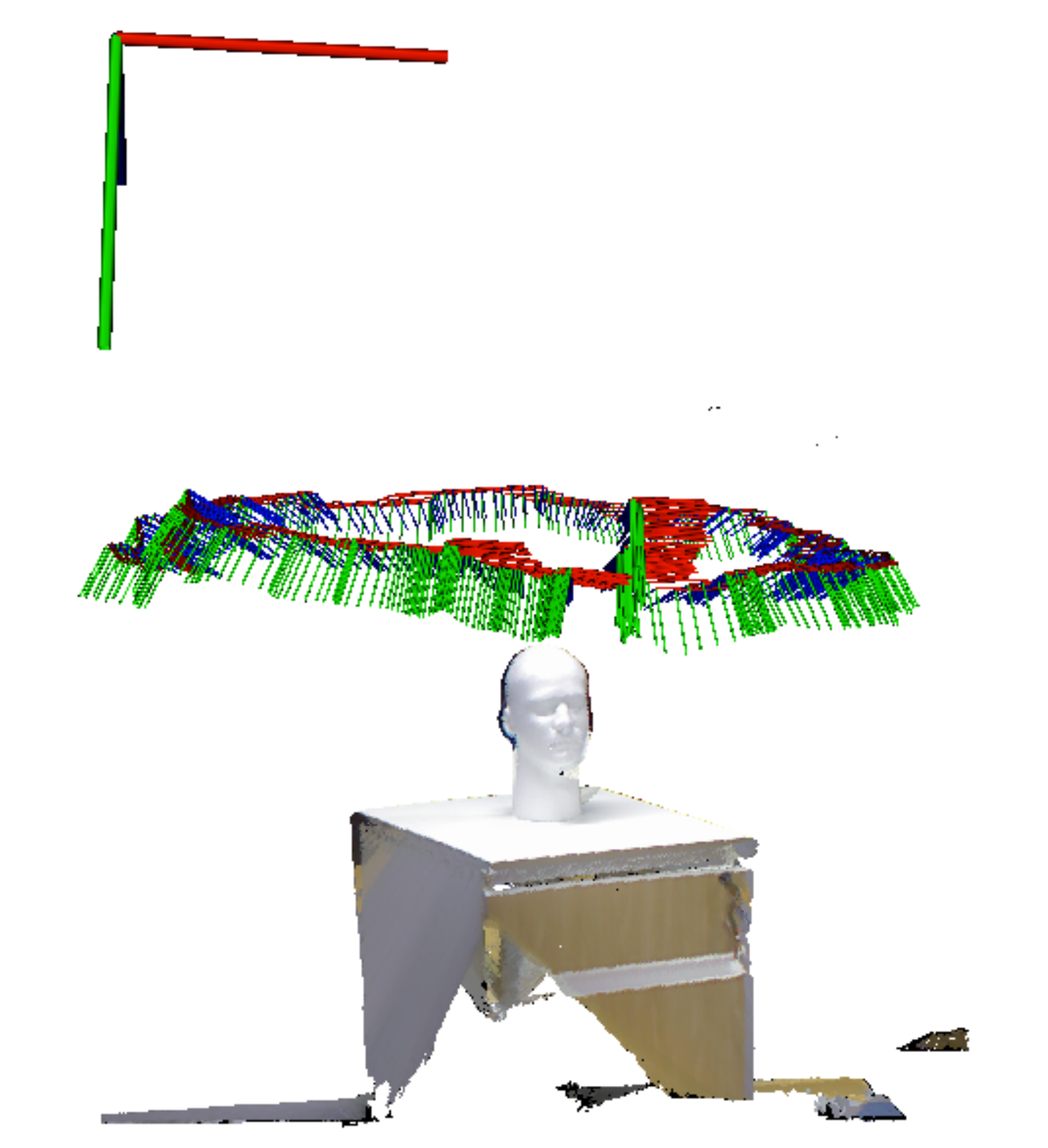}} \qquad %
	\subfloat[Top view of selected table plane and sensor poses]
	{\includegraphics[height=3cm]{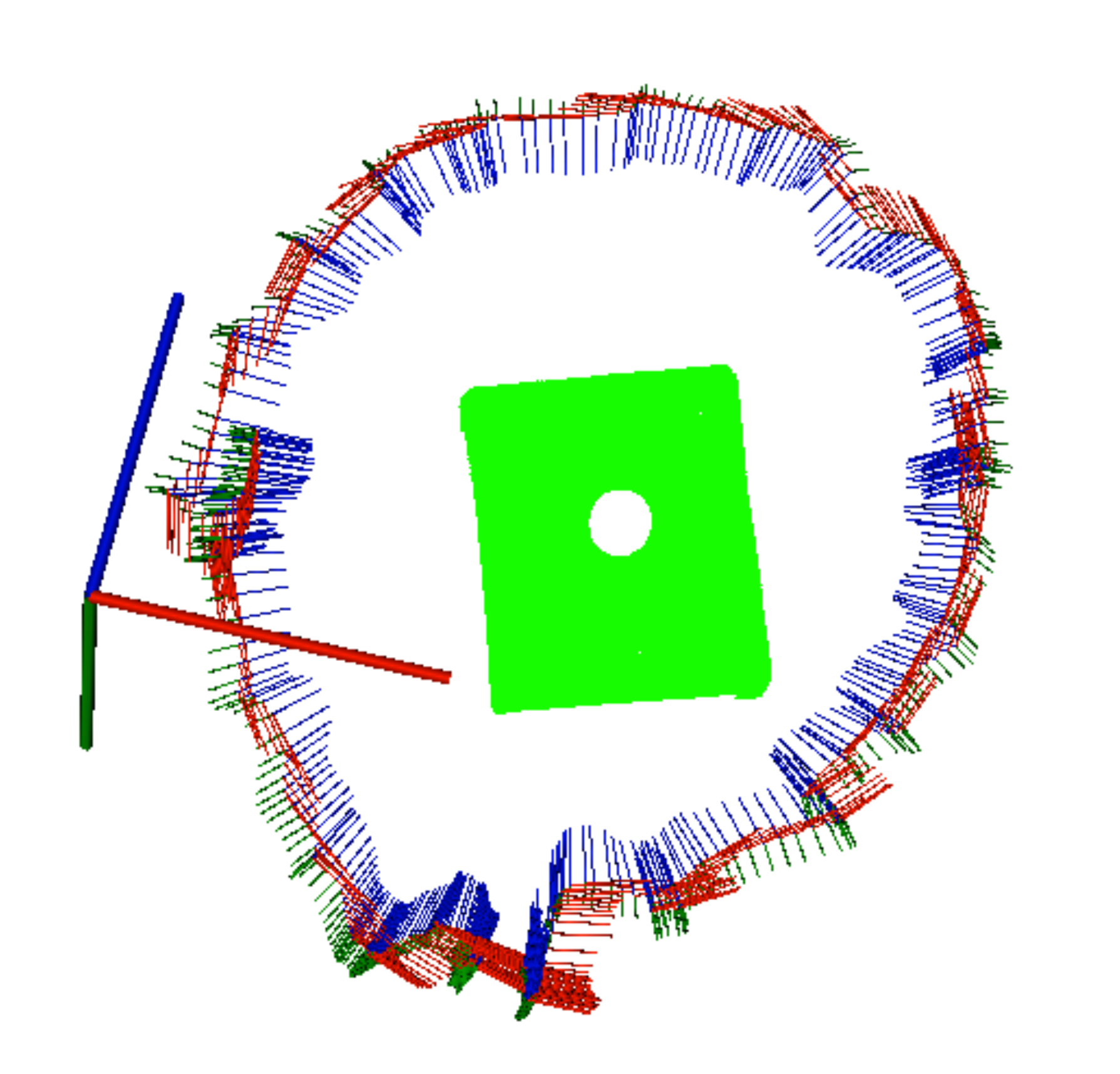}} \qquad
	\subfloat[Side view of selected table plane and sensor poses]
	{\includegraphics[height=3cm]{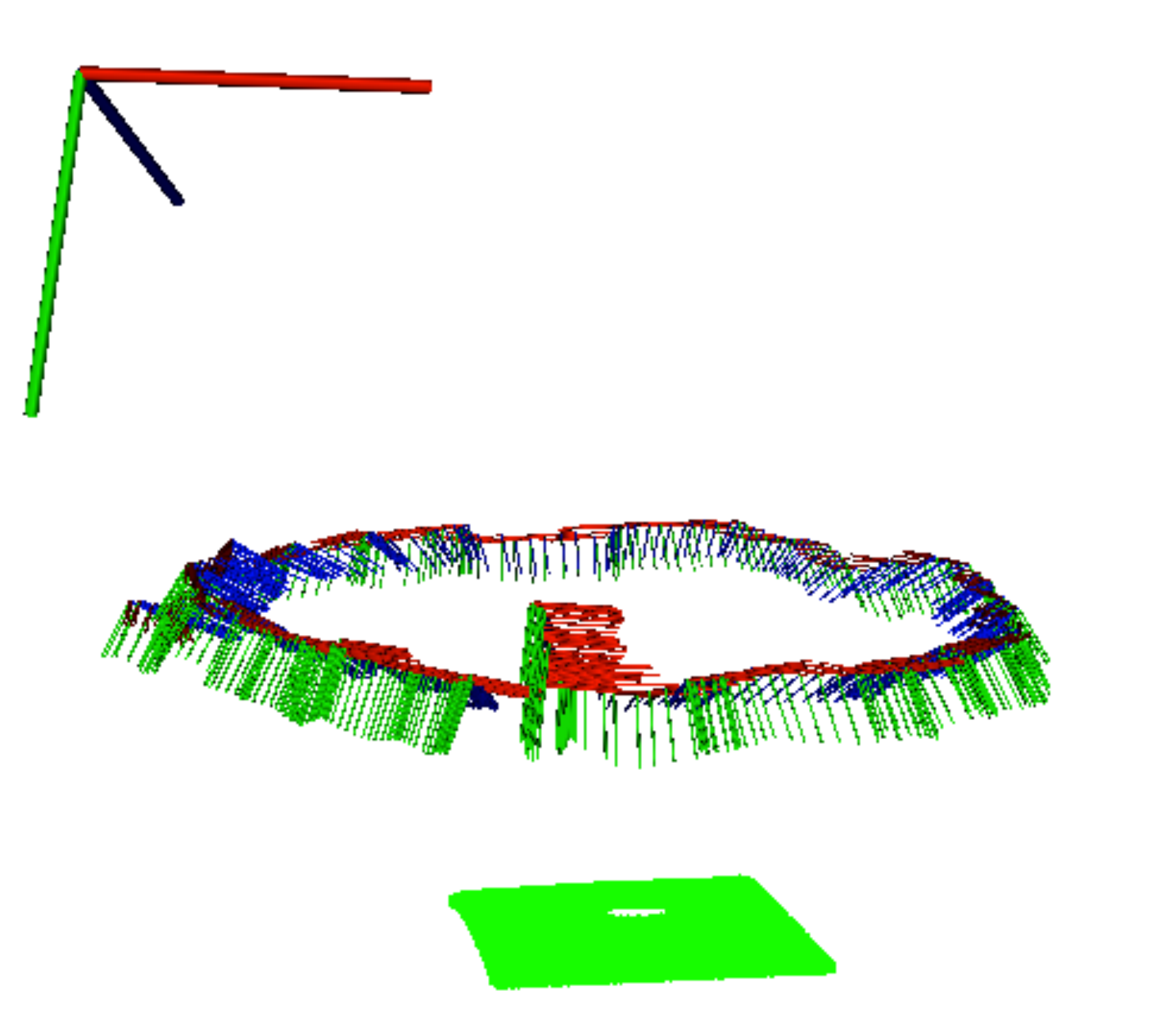}}
	\caption{Result of applying RANSAC based planar model fitting to the scene. (a) Full 3D reconstruction (b) Top view and (c) Side view of the table-top plane in green. Sensor poses $c_i$ are represented by the multiple coordinate frames.}
	\label{fig:planeposes}
\end{figure}

\subsubsection{Selection for Standard Human Head Model}
\label{ssec:objectseg}

The first step for selecting the reconstructed head model \emph{(with the assumption that it lies on a table)} is to find the table top area where the standard human head model is located. We use a Random Sample Consensus (RANSAC)-based method to iteratively estimate parameters of the mathematical model of a plane from a set of 3D points of the scene \cite{Fischler-ACM81}. The mathematical model of a plane is specified in the Hessian normal form as follows:

\begin{equation}
ax + by + cz + d = 0
\label{eq:plane}
\end{equation}
where $a,b,c$ are the normalized coefficients of the $x,y,z$ coordinates of the plane's normal and $d$ is the Hessian component of the plane's equation. The largest fitted plane is selected from the point cloud, this plane represents the object-supporting surface (i.e., table or counter) of the scene. Now that the plane of the table-top has been identified, we are interested in extracting the set of points that lie on top of this plane and below the maximum z-value of the set of sensor poses $C$, with respect to the table plane. (Fig. \ref{fig:planeposes})

In order to extract the points ``on top'' of the table without removing any useful information, we transform the plane so that its surface normal $n_{i}$ directions are parallel to the z-axis and the plane is orthogonal to the z-axis and parallel to the x-y plane of the world coordinate system.

\begin{algorithm}[ht]
	\small
	\renewcommand{\algorithmicrequire}{\textbf{Input:}}
	\renewcommand{\algorithmicensure}{\textbf{Output:}}
	\caption{Selection for Standard Human Head Model}
	\label{alg:frame}
	\begin{algorithmic}
		\REQUIRE T (rigid transformation), $P_{plane}$(point cloud of the selected plane), $P_{full}$(full point cloud of reconstruction), $C$(tracked sensor poses from reconstruction)
		\ENSURE  $P_{object}$ (point cloud representing the object on top of the table)
		\STATE $P_{plane}^{*} = T \cdot P_{plane}$
		\STATE $P_{full}^{*} = T \cdot P_{full}$
		\STATE $C^{*} = T \cdot C$
		\STATE $3DPrism \leftarrow construct3DPrism(P_{plane}^{*})$
		\STATE $P_{object}^{*} \leftarrow extractPointswithinPrism(C^{*},P_{full}^{*},3DPrism)$
		\STATE $P_{object}=T^{-1}\cdot P_{plane}^{*}$
	\end{algorithmic}
\end{algorithm}

\begin{figure}[t]
	\centering
	\subfloat[3D prism for selection.] 
	{\includegraphics[height=3cm]{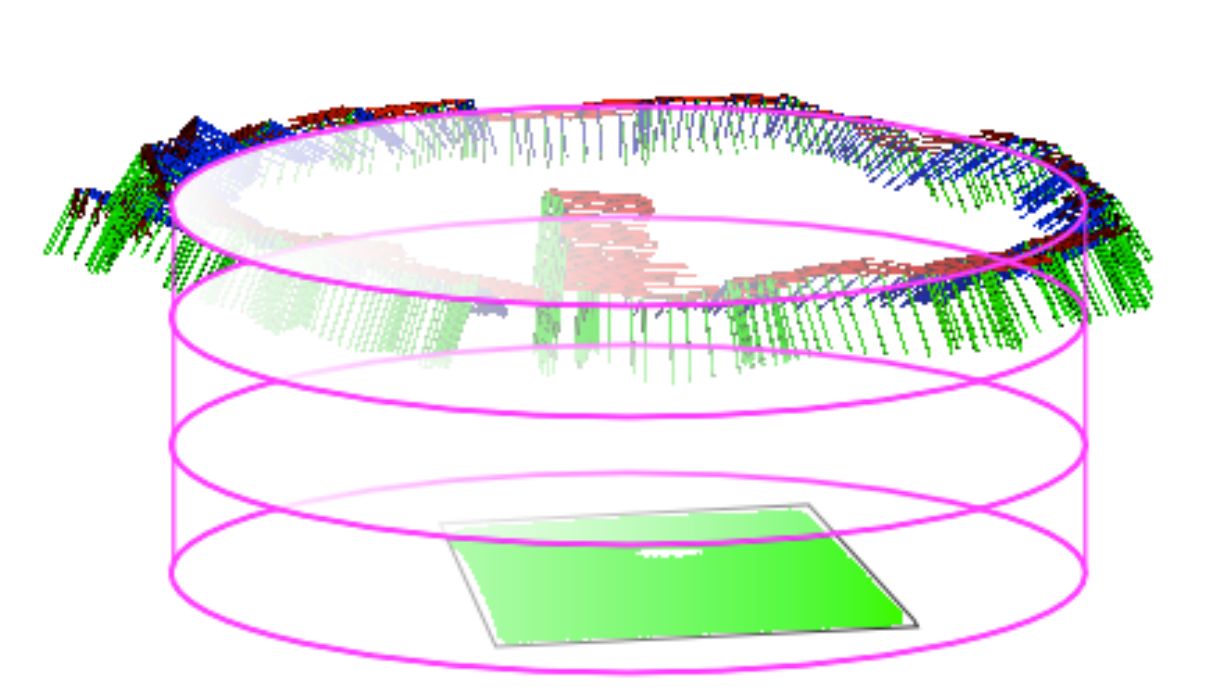}} \qquad %
	\subfloat[Segmented 3D model (blue)]
	{\includegraphics[height=3cm]{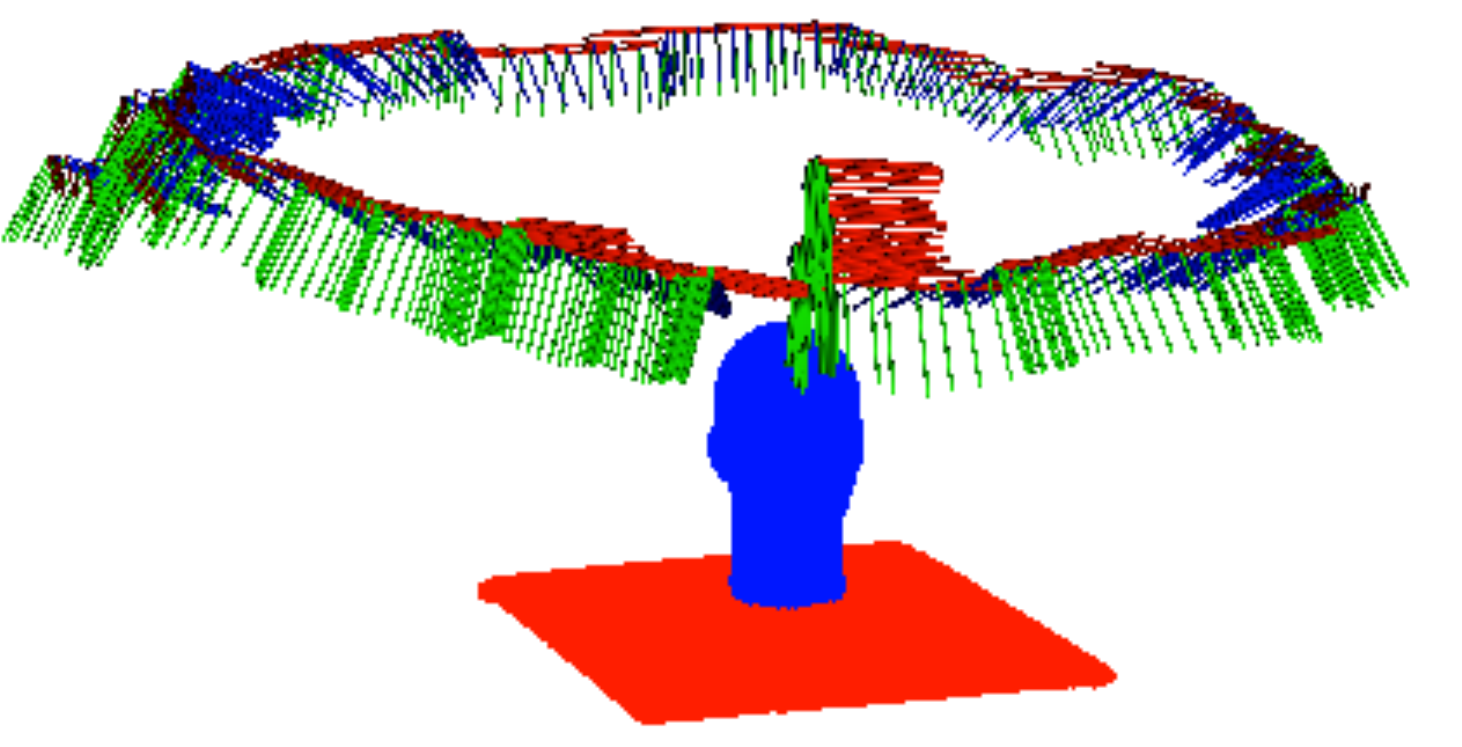}}
	\caption{Selection procedure for standard human head model. (a) Constructed 3D prism which represents the scanned head model (b) 3D model after selection}
	\label{fig:hull}
\end{figure}

Finally, we construct a homogeneous transformation matrix $T_p=[R,t]$, where $t=[0,0,0]^{T}$. The procedure to extract the object from the full 3D reconstruction is listed in Algorithm \ref{alg:frame}. Specifically, $P_{plane}$, $P_{full}$ and $C$ are transformed by $T$ so that the plane is orthogonal to the z-axis. Then we create a 3D prism between the convex hull of the $P_{plane}$ and the convex hull of the loop generated from the sensor poses $C$. As seen in Fig. \ref{fig:hull}, the face of the 3D prism is contracted from the convex hull of the shape of the loop of sensor poses projected to the plane. This shape is then extruded in the negative Z direction until it crosses the table plane. The points within this 3D bounding prism are extracted and transformed back to the original world coordinate system, resulting in a point cloud containing only the object on the table top.

%

\subsubsection{Selection for Real Human Head}

Since human head models are usually produced in the form of human busts, which contains a little part of human upper body, to increase stability, we also propose an automatic selection for real human bust as well. When scanning a human to print a bust, we generally tend to concentrate on the head, even though the human is standing or sitting down. There is no ground plane in either case. The only spatial knowledge we have about the scan is the sensor poses and that the human is sitting or standing upright. However, in order to create our 3D prism for selection, as mentioned in the last section, we create a virtual plane on top of the subject's head. The procedure is listed in Algorithm \ref{alg:humanseg}.

\begin{algorithm}[ht]
	\small
	\renewcommand{\algorithmicrequire}{\textbf{Input:}}
	\renewcommand{\algorithmicensure}{\textbf{Output:}}
	\caption{Selection for Real Human Head}
	\label{alg:humanseg}
	\begin{algorithmic}
		\REQUIRE  $P_{full}$(full point cloud of reconstruction), $C$(Tracked sensor poses from reconstruction)
		\ENSURE  $P_{human}$ (point cloud representing the human bust)
		\STATE $C_{centroid} = compute3DCentroid(C)$
		\STATE $Head_{top} = nearestNeighbors(C,k,P_{full})$
		\STATE $P_{plane} = fitplane(Head_{top})$
		\STATE $T = findPlaneTransform(P_{plane})$
		\STATE $P_{plane}^{*} = T \cdot P_{plane}$
		\STATE $P_{full}^{*} = T \cdot P_{full}$
		\STATE $C^{*} = T \cdot C$
		\STATE $3DPrism \leftarrow construct3DPrism(C^{*},P_{plane}^{*}, offset_{head})$
		\STATE $P_{human}^{*} \leftarrow extractPointswithinPrism(P_{full}^{*},3DPrism)$
		\STATE $P_{human}=T^{-1}\cdot P_{plane}^{*}$
	\end{algorithmic}
\end{algorithm}

%

\begin{figure}[htbp]
	\centering
	\subfloat[$Head_{top}$, $P_{plane}$ and $C_{centroid}$] 
	{\includegraphics[width=5cm]{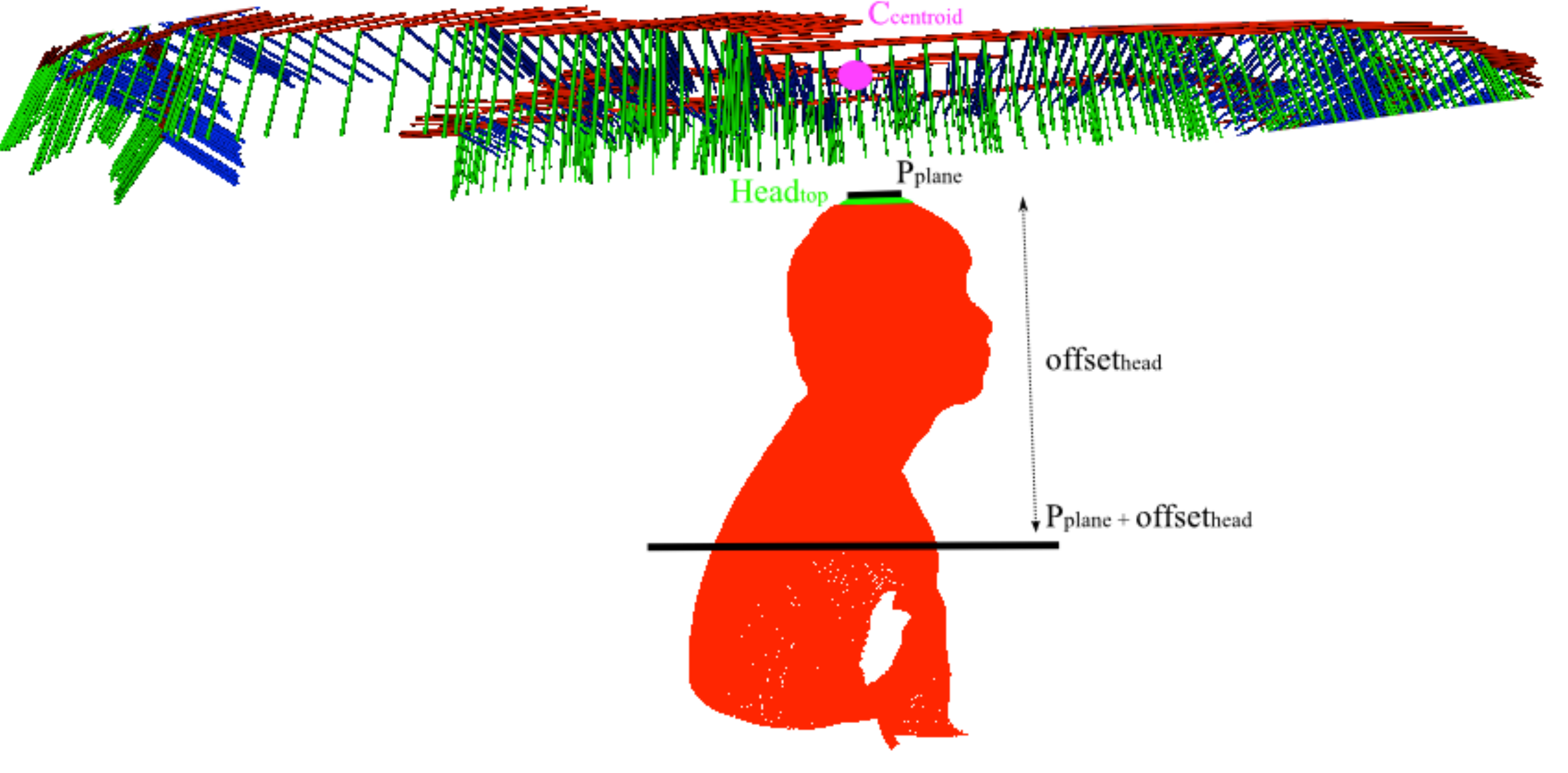}} \qquad %
	\subfloat[3D prism for selection.]
	{\includegraphics[width=5cm]{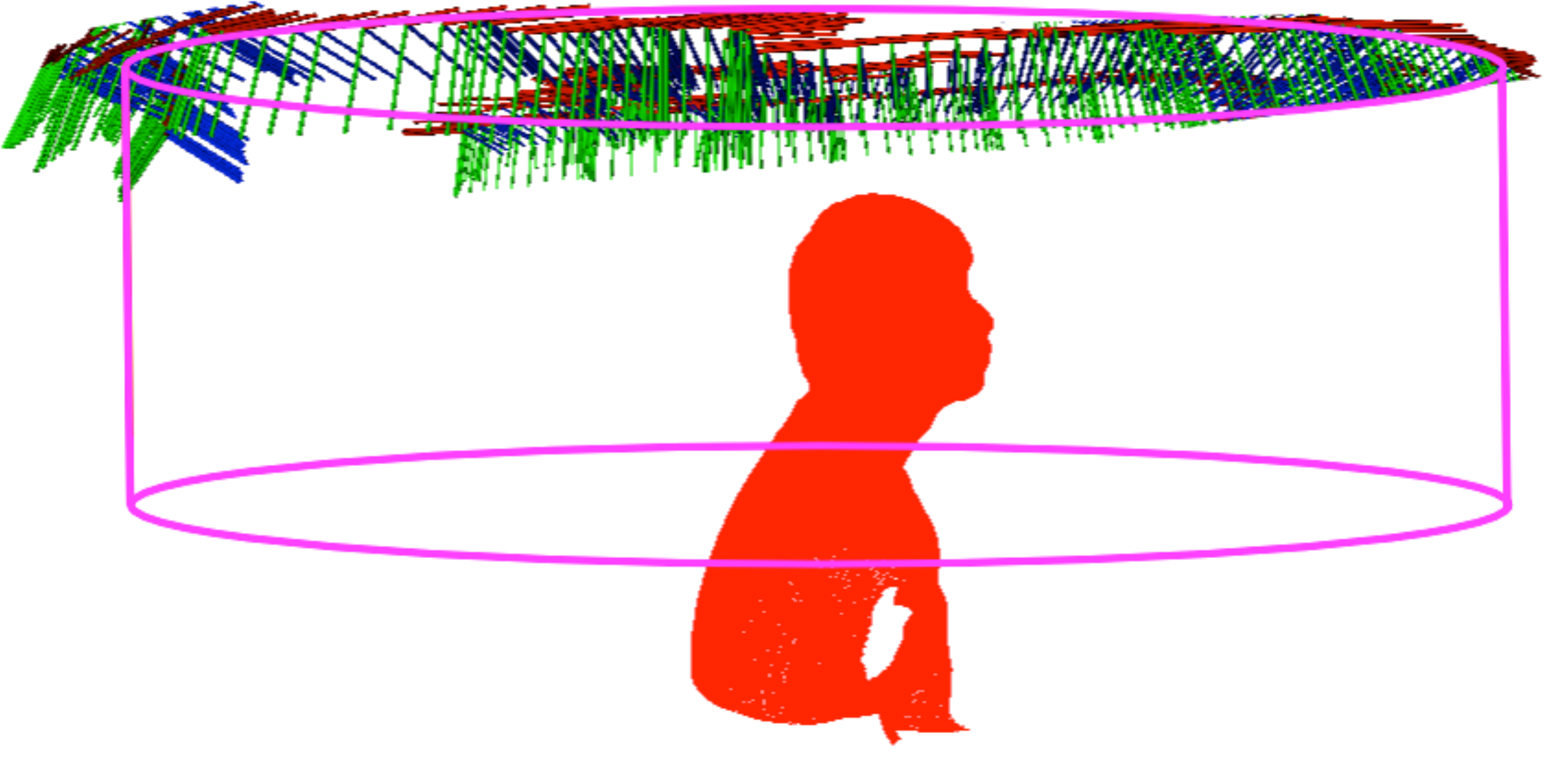}}
	\caption{Selection for real human head. (a) Side view of top head points ($Head_{top}$) virtual plane ($P_{plane}$), camera pose centroid and ($C_{centroid}$) (b) constructed 3D prism which represents the human head.}\label{fig:humanprism}
\end{figure}

As we can see, the difference between Algorithm 2 and Algorithm 1 is only the first four lines, where we create a virtual plane in order to construct the 3D prism for selection. Initially, we compute the 3D centroid of the sensor poses. Then, we find the $k$ nearest neighbors of the centroid with respect to the human reconstruction. A reasonable value for $k$,  can vary depending on the resolution of the reconstruction and the total number of points. These points ($head_{top}$) represent the closest set of points from the top of the head of the human subject to the centroid of the sensor poses. Once $head_{top}$ is obtained, we search for a planar model as in Section \ref{ssec:objectseg} and further estimate the planar transformation matrix $T$ for $P_{plane}$. The resulting 3D prism and human head selection can be seen in Fig. \ref{fig:humanprism}.

\subsection{Scaling Post-Processing}

\begin{figure}[ht]
	\centering
	\subfloat[Real spatial dimensions of reconstructed model.] 
	{\includegraphics[width=4cm]{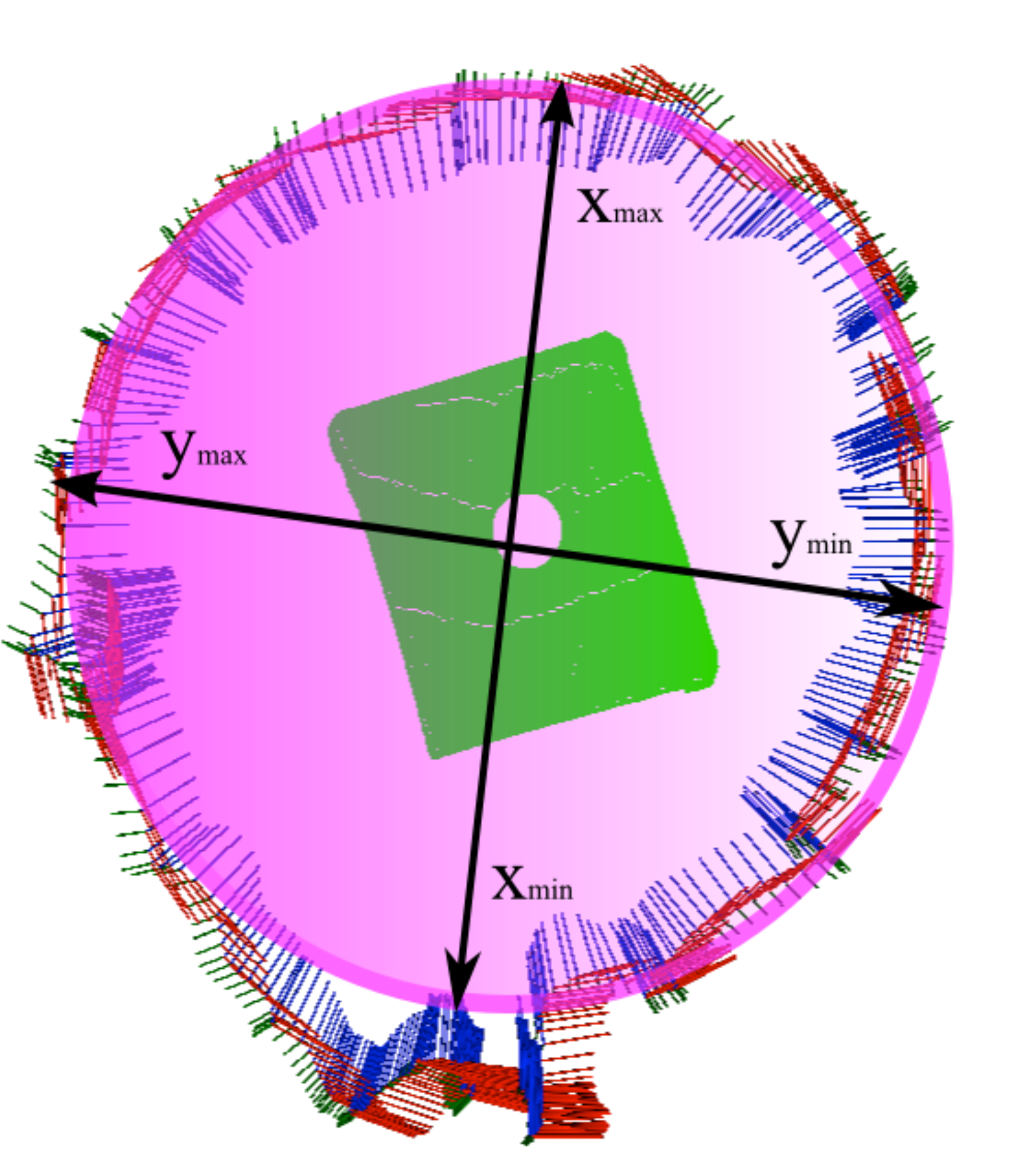}} \qquad %
	\subfloat[Scaled spatial dimensions to printer size.]
	{\includegraphics[width=4cm]{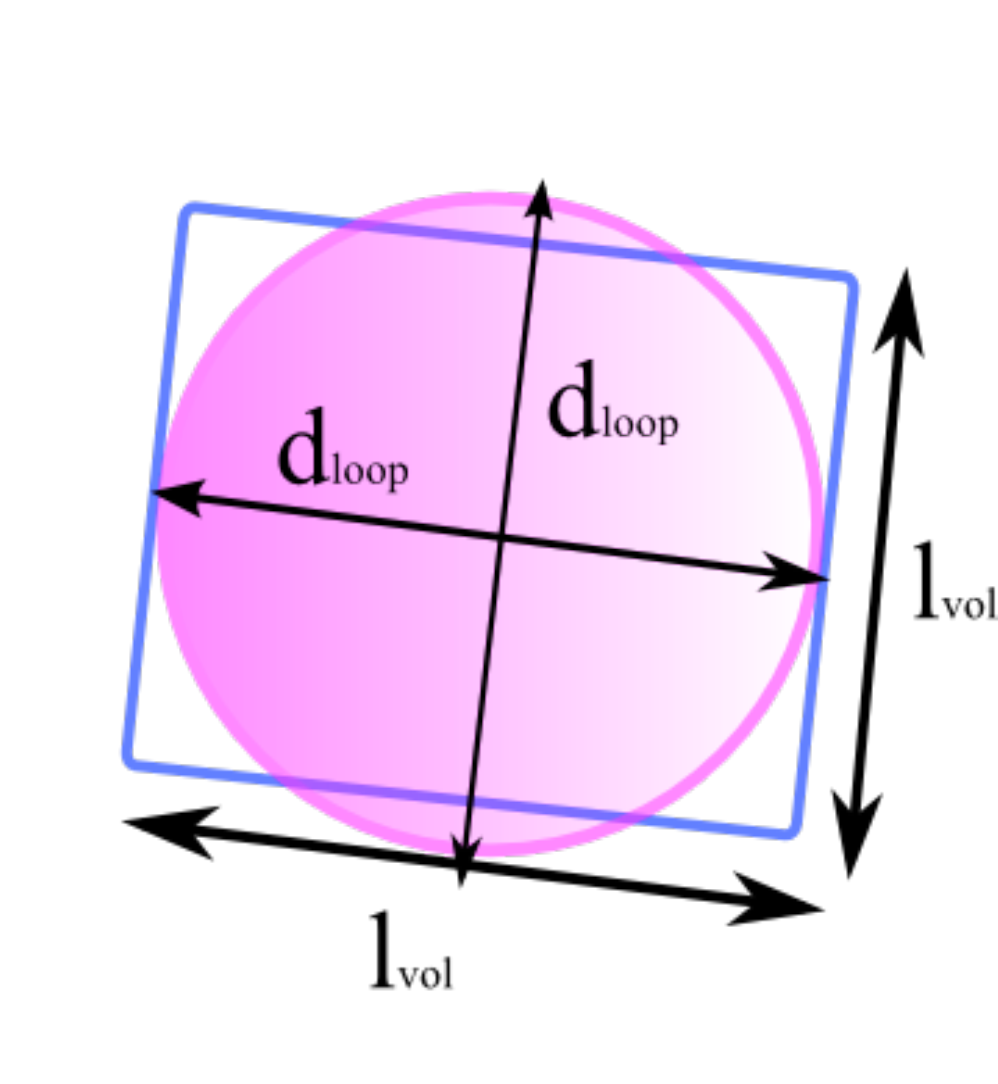}}
	\caption{Automatic scaling of the reconstructed model to the volume of a 3D printer. (a) Real dimensions (b) scaled dimensions.}\label{fig:scaling}
\end{figure}

In order to have a direct scaling of the world to the 3D volume of a 3D printer, we create a circle which represents the closed loop of the measurements on the $X$-$Y$ plane, namely, the approximate camera pose loop. The diameter of this approximate loop ($d_{loop}$) is obtained by computing the maximum distance between axis from the sensor poses:
\begin{equation}
d_{loop} =\arg\max(||x_{min} - x_{max}||, ||y_{min} - y_{max}||)
\end{equation}
Since the maximum possible diameter of the scaled world in the printer's volume is the length ($l_{vol}$) of the face of the model base, the scaling factor ($sf$) is computed as follows:
\begin{equation}
sf = l_{vol}/d_{loop};
\end{equation}
The final selected reconstructed model is then scaled by $sf$. Fig. \ref{fig:scaling} shows the automatic scaling of the reconstructed model to the volume of a 3D printer.

\subsection{3D Printing}

\begin{figure}[htbp]
	\centering
	\subfloat[3D part for printing of an inanimate standard head model.] 
	{\includegraphics[height=4cm]{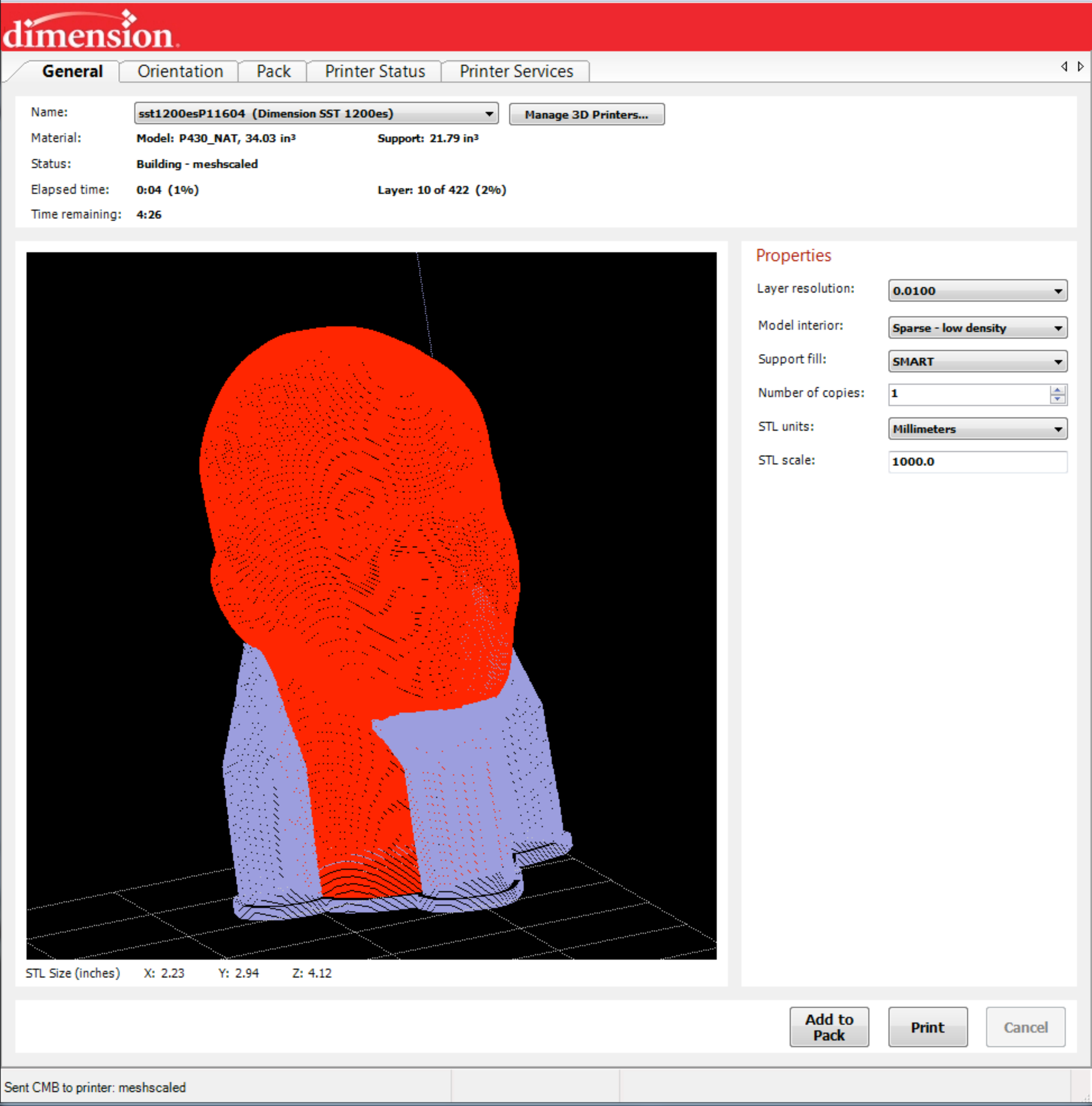}} \qquad %
	\subfloat[3D part for printing of a human head.]
	{\includegraphics[height=4cm]{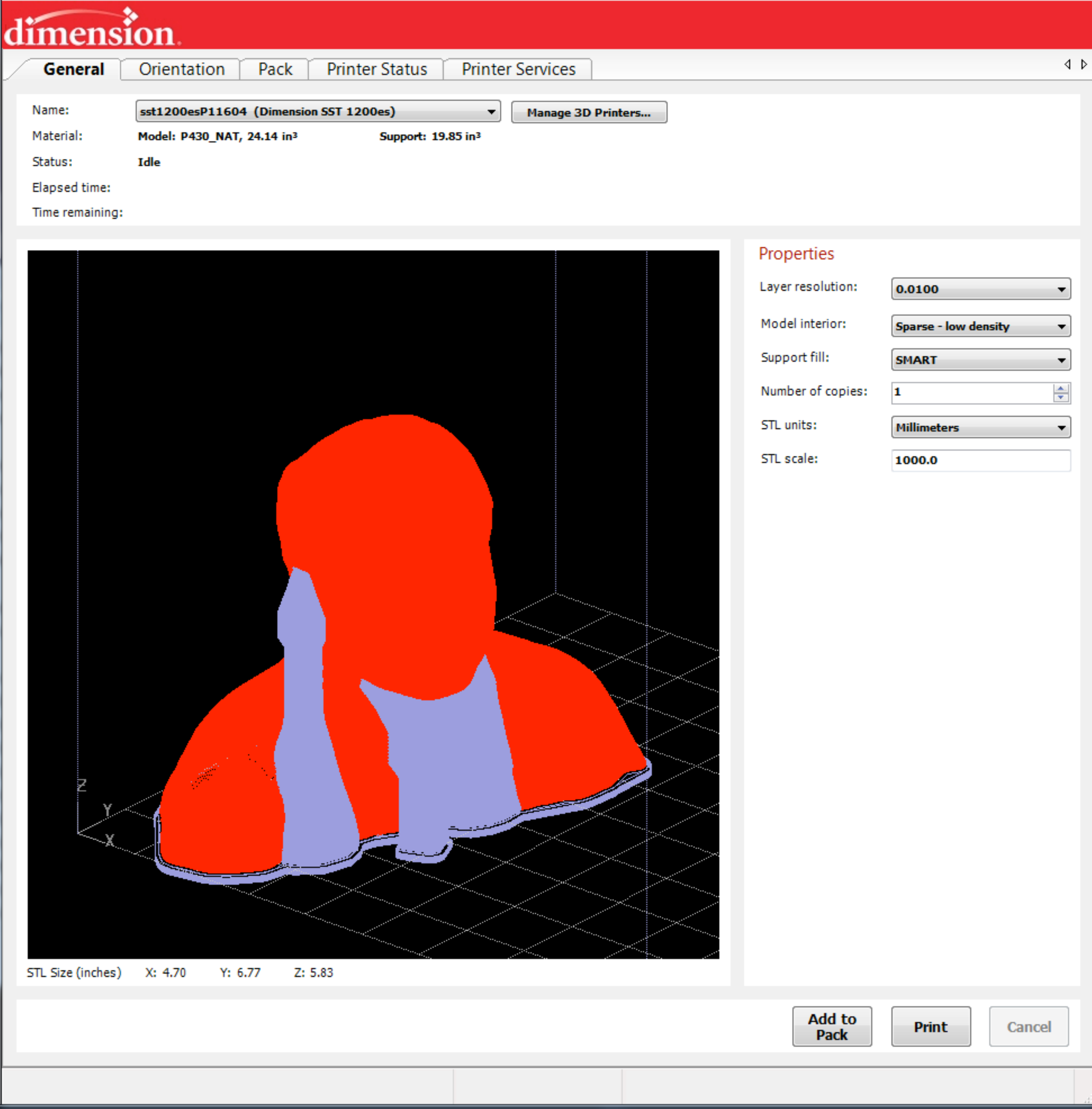}}
	\caption{Generated layers and support material for the final 3D model. (a) Standard human head model (b) Real human head. }\label{fig:printersoftware}
\end{figure}

After applying selection and scaling to the reconstructed model, we generate a printable polygon mesh. The system then automatically uploads the resulting .stl file through WiFi to the machine connected to the 3D printer. Once a model of the printer has been imported to the modeling software (Fig. \ref{fig:printersoftware}), the layers as well as the necessary support material are computed. However, even if we automatize this procedure when sending the ready part to the printer a confirmation button has to be pushed and a new model base has to be loaded. If the printer manufacturers provide the option of having a remote interface, we can fully accomplish the automatic pipeline.


\section{Implementation and Results}
\label{sec:ExpResults}

\subsection{Implementation}

The implementation of our proposed system was conducted with the following hardware and software:

\begin{itemize}
	
	\item To enhance the mobility of our system, we used DFRobot's HCR Mobile Robot to carry the composite sensor, its external battery, a tripod, and a tablet (Microsoft Surface) which realized all computational work. PID control was adopted to make the robot circulate around the subject with approximate 1-meter radius. In addition, Kinect v2's power adaptor was replace by an external battery to avoid cable limitation. 
	
	\item To acquire and fuse depth data from Kinect v2 (with depth image resolution 512$\times$424), we modified KinectFusion and open-source Point Cloud Library (PCL) \cite{Rusu_ICRA2011_PCL,Tist} on the tablet as introduced before.
	
	\item The 3D printer we adopted is Stratasys' Dimension 1200es, a professional 3D printer with the build size 254$\times$254$\times$305mm and layer thickness 0.25mm. Its printing technology is Fused Deposition Modeling (FDM), which is based on material extrusion. The accuracy of this 3D printer is $\pm0.127mm$, which is acceptable when considering the size of the head model.
	
	\item The software we used to compute two 3D models' Hausdorff distances is MeshLab\footnote{MeshLab. http://meshlab.sourceforge.net/} version 1.3.3, an open-source 3D tool supported by the 3D-CoForm project. MeshLab provides the function to calculate and visualize Hausdorff distances. 
	
\end{itemize}

The complete flow of our implementation, with both standard human head model and real human subject, are shown in Fig. \ref{fig:flowchart}. In addition, more detailed experimental results are shown in the following subsections.

\subsection{Computational Cost}

Currently, the reconstruction algorithm runs almost real-time (i.e., 15 frames per second) and the post-processing steps take a few minutes depending on the subject and the scanning quality. The required time for 3D printing also depends on the complexity and the size of the models, as well as the printer itself. For example, in our implementation, the standard head model took approximately 7 hours to print, whereas the head model from real head takes approximately 12 hours to print.

%

\subsection{Accuracy of Reconstructed Head}

To test the accuracy of our results, we compared the 3D models generated by our proposed system with the ones reconstructed by FastSCAN\footnote{FastSCAN. \url{http://polhemus.com/scanning-digitizing/fastscan-cobra-ci/}} and Cyberware\footnote{Cyberware. \url{http://cyberware.com/products/scanners/px.html}} commercial laser scanning systems on both printed standard head model and real human subject separately. All the scanned 3D data is published on \textit{The MCRLab 3D Human Head Scanning Repository \footnote{The MCRLab 3D Human Head Scanning Repository. \url{https://2d38791be59e3ad62ca59c967554ea6389b91bf9.googledrive.com/host/0B4TnftTwy3ThVXQ3R2lEb0I1aGs/3D_Head_Repository.html}}}. We computed the geometric differences between the 3D model data acquired by different scanners. Our numerical evaluation is based on computing the approximate error between two triangular meshes representing the same surface or object ($M_1 \leftrightarrow M_2$), as introduced by Cignoni et. al. \cite{Metro}. The approximation error is defined with the two-sided Hausdorff distance $d_H$ \cite{Hausdorf}, which is the maximum value from $M_1 \rightarrow M_2$ and $M_2 \rightarrow M_1$ in the Euclidean space, as follows:
\begin{equation}
	\begin{split}
	 &d_{H}\left(M_{1},M_{2}\right)\\
	 &=  \max\left\{ \sup_{m_{1}\in M_{1}}\left(\inf_{m_{2}\in M_{2}}d\left(m_{1},m_{2}\right)\right)\right.\nonumber
	\left.,\sup_{m_{2}\in M_{2}}\left(\inf_{m_{1}\in M_{1}}d\left(m_{1},m_{2}\right)\right)\right\}\\
	\end{split}
\end{equation}
where $\operatorname{sup}$ represents the supremum, $\operatorname{inf}$ is the infimum, $m=(x,y,z)$ is the 3D vertex points of the corresponding triangular mesh and $d$ is the Euclidean distance between two points in Euclidean space $E^3$. To provide comparative results between the ground-truth model ($M_{gt}$) and the laser scanned model ($M_f$), as well as the Kinect v2 scanned mode ($M_k$), we take the full set of vertex points (68k) from the ground truth model and search for the closest point on the scanned models to compute error metrics based on the Hausdorff distance $d_H$. The unit to represent Hausdorff distance in MeshLab is Hausdorff Distance unit (HDu), represented as color bar in Fig. \ref{fig:resultsviz} and \ref{fig:Real_all}.  

\subsubsection{Standard Human Head Model}
\begin{figure}[ht]
	\centering
	\includegraphics[width=.8\linewidth]{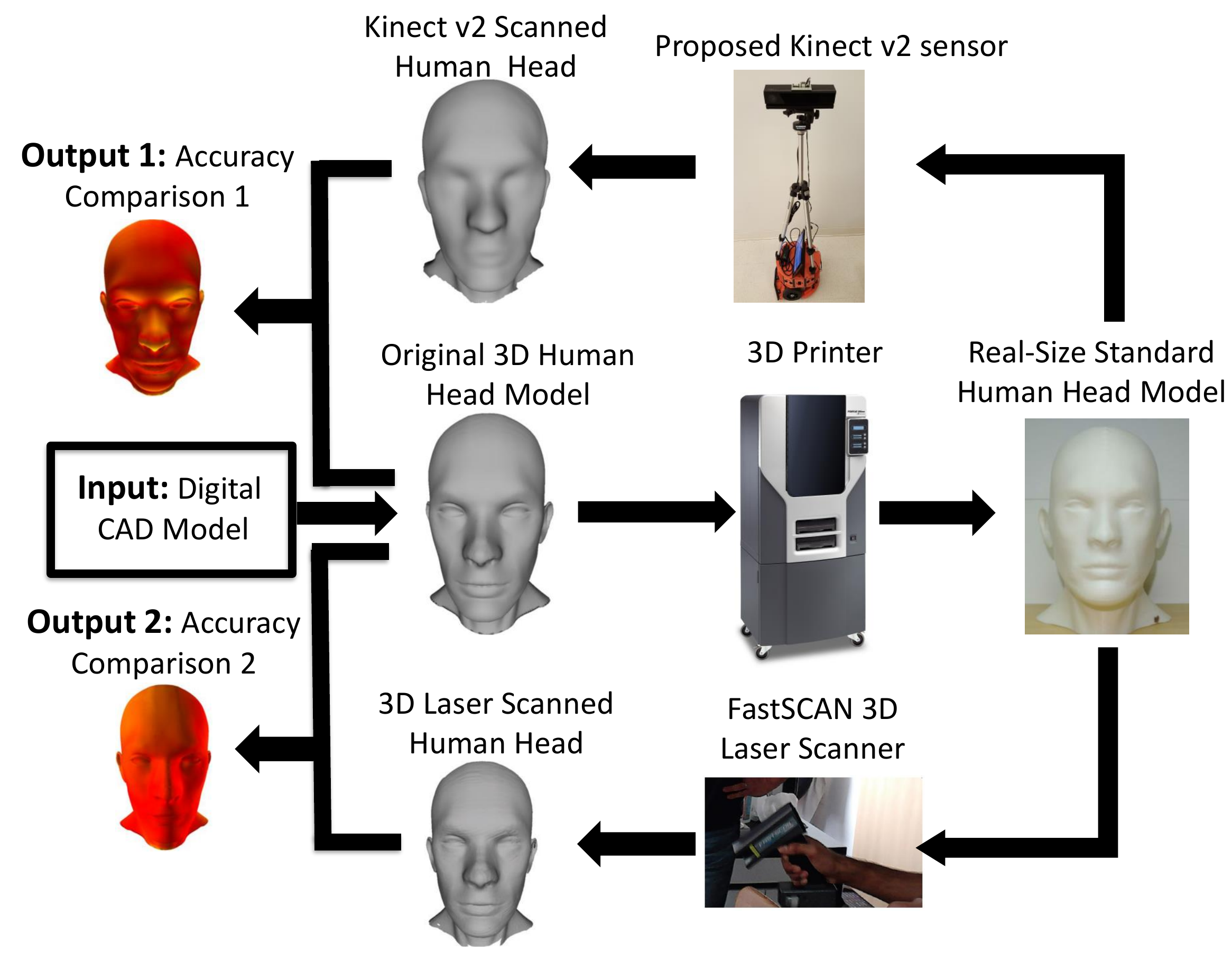}
	\caption{Experimental setup for the evaluation of standard head model scanning.}
	\label{fig:evaluation}
\end{figure}

To validate the capabilities of our proposed scanning system, a visualized 3D CAD model of a standard human head model is prototyped, 3D printed and scanned separately by our proposed scanning system and by a commercial handheld 3D laser scanner FastSCAN. Furthermore, we computed the geometric differences (represented by Hausdorff distances) between the two scanned 3D models and the ground-truth model separately, as illustrated in Fig. \ref{fig:evaluation}.

In Table \ref{tab:results}, we present the accuracy results for each scanned model. Since we have the detail model and the printed size of the original ground-truth model, we translate Hausdorff Distance unit (HDu) from MeshLab to real-word distance.  Each column corresponds to the scanning system that was used (first column: FastSCAN, second column: proposed system). For each scanned model, we compute the mean, maximum and RMS (Root Mean Square) translated from the Hausdorff distances between all the points (first multi-row). We also present the error with respect to the diagonal of the bounding box of the mesh (second multi-row), which, to the human eye, is a more understandable error.

\begin{table}[ht]
	\caption{Comparative results with the standard human model}
	\label{tab:results}
	\centering
	\begin{tabular}{ llcc }
		\hline%
		\multicolumn {2}{ c }{ Error Metrics } &  FastSCAN  &   Proposed\\
		&  &  System &  System\\\hline%
		Hausdorff & mean & 0.1868 & 0.5443\\\cline {2 -4}
		Distance [cm] & max & 0.6177 & 1.5026\\\cline{2-4}
		& RMS & 0.2257& 0.7580\\\hline
		Bbox & mean & 0.4283 & 1.2853\\\cline {2 -4}
		Diagonal [\%]  & max & 1.4168 & 7.1472\\\cline{2-4}
		& RMS & 0.5177 & 1.7901\\\hline
	\end{tabular}
\end{table}

\begin{figure*}[t]
	\centering
	\includegraphics[width=.9\linewidth]{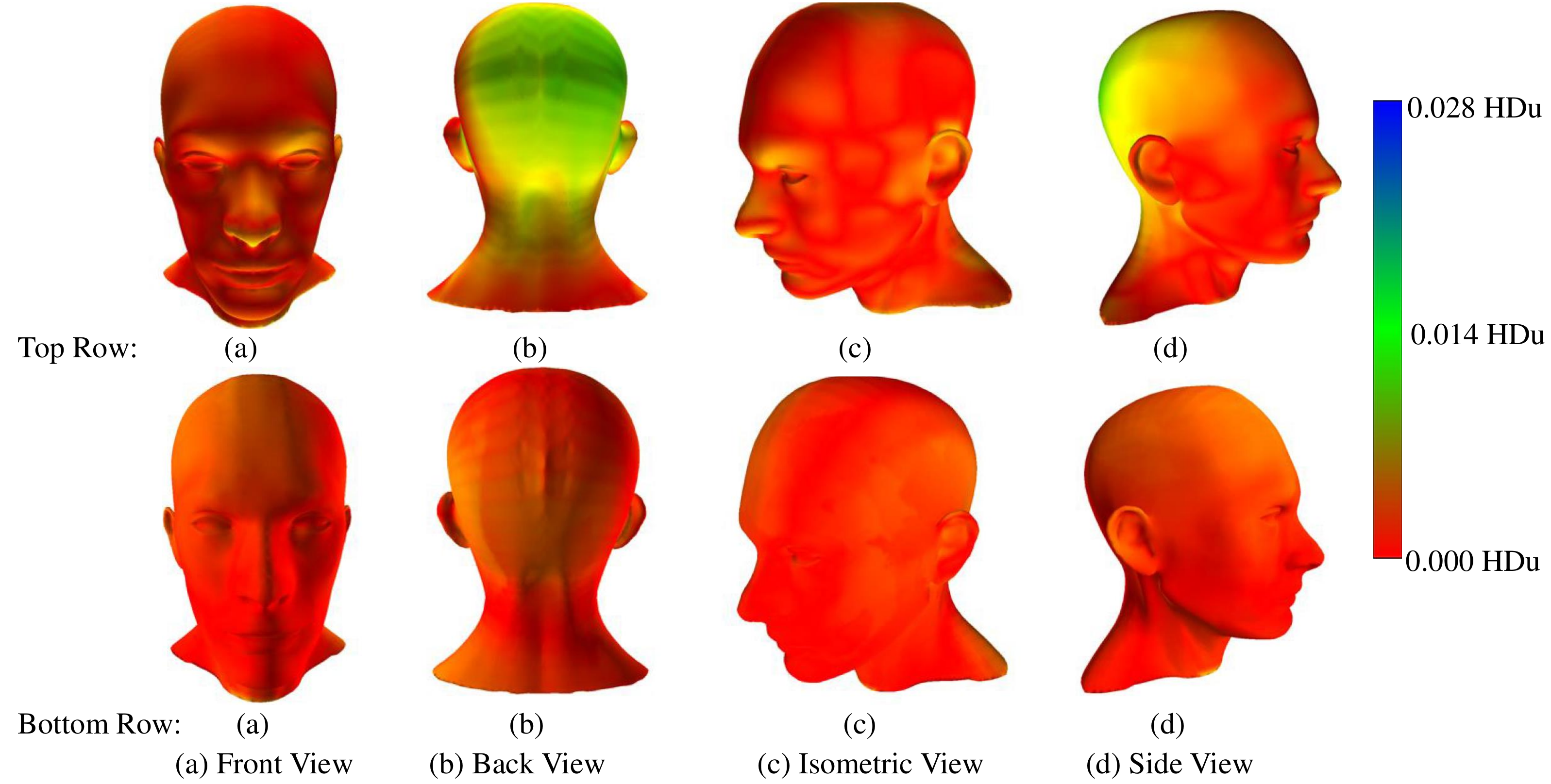}
	\caption{Visualization of Hausdorff distance between scanned models and the ground-truth model separately. Top Row: Proposed system result. Bottom Row: FastSCAN scanner result.}
	\label{fig:resultsviz}
\end{figure*}

These errors are visualized in Fig. \ref{fig:resultsviz}. Same as we predicted, the FastSCAN laser scanning result showed a higher degree of accuracy than our proposed Kinect v2 sensor based scanning system. However, the maximum Hausdorff distance error of our proposed system is only two-fold that of the laser scanner, the mean and RMS errors stay within the same decimal range. Furthermore, if we analyze the values with respect to the bounding box diagonal we show an error of less than 2\%, which is acceptable for the types of applications for which the scanner is intended. If we take a closer look at the error visualization, we can identify where our proposed system exceeds the average distance compared to the original mesh. In facial areas such as the nose and eyes, the  Kinect v2 scanner fails due to the proximity of the sensor (which has to be approximately 0.5m, with depth image resolution 512$\times$424), this produces the lack of detail in these areas. Another problematic area is behind the ears, also due to the proximity-resolution of the sensor. Finally, the most error prone area is the back of the head. In this case, ICP may get lost when evaluating such similar surface patches. The FastSCAN scanner on the other hand, has no problem with these previously mentioned areas. One of the reasons is that the proximity of the sensor is approximately 0.1m, thus it achieves much higher resolution and can be swept many time within corners, such as behind the ears to achieve an accurate model. Based on these facts and the price difference, result from our proposed system is highly acceptable. 

\subsubsection{Real Human Head}

\begin{figure*}[htbp]
	\centering
	\includegraphics[width=.94\linewidth]{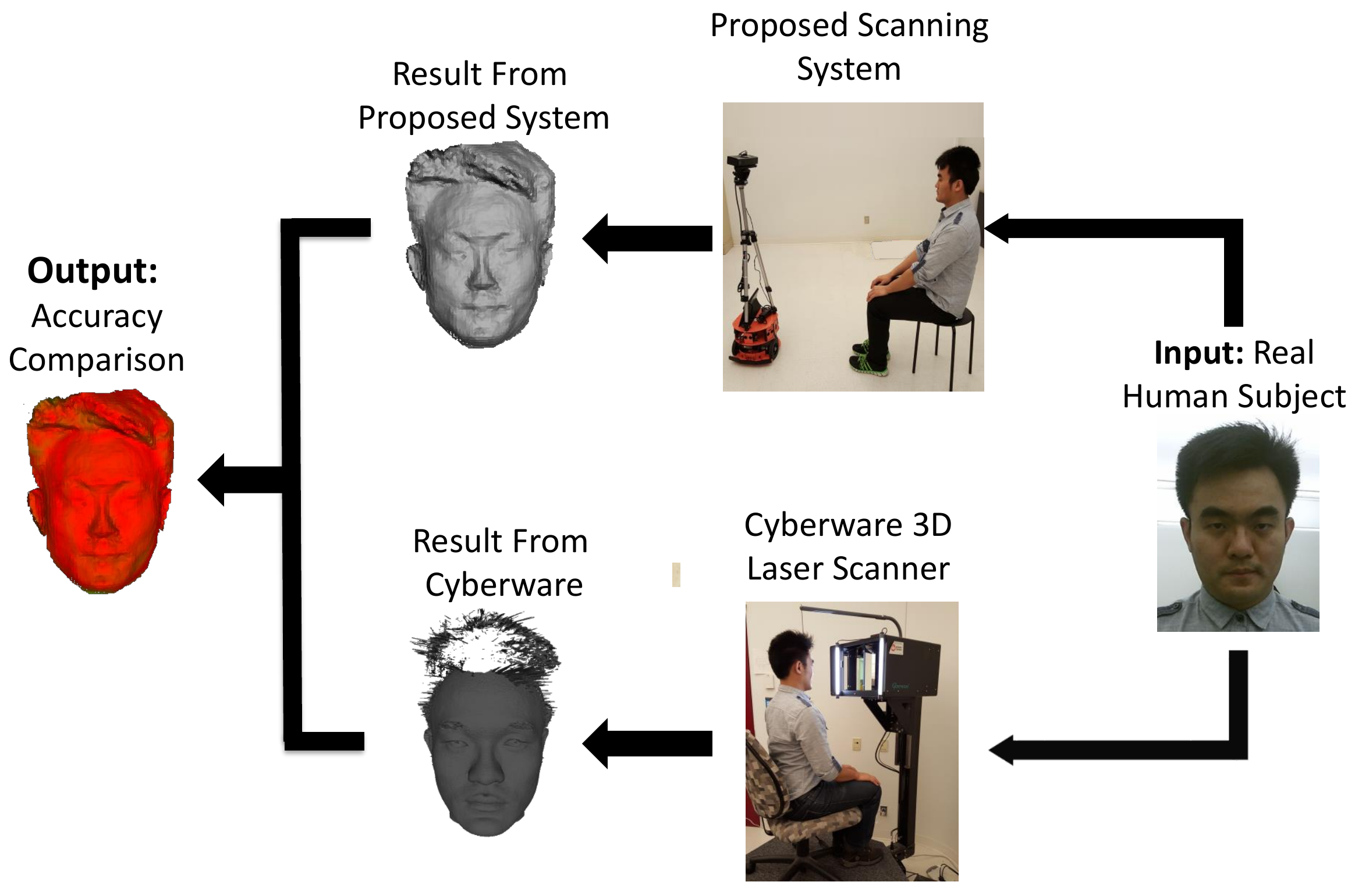}
	\caption{Experimental setup for the evaluation of real human scanning.}
	\label{fig:Real_evaluation}
\end{figure*}

Previous comparison was implemented on a 3D printed inanimate head made with only one kind of homogenous texture-less material. To further validate our experimental results, we compared scanned models from our proposed system with the ones from commercial Cyberware laser scanning system on real-human subjects. The experimental setup is illustrated in Fig. \ref{fig:Real_evaluation}. We scanned real human subjects with both proposed system and Cyberware separately, and then calculated the Hausdorff distance between the two acquired models.

\begin{figure*}[htbp]
	\centering
	\includegraphics[width=.9\linewidth]{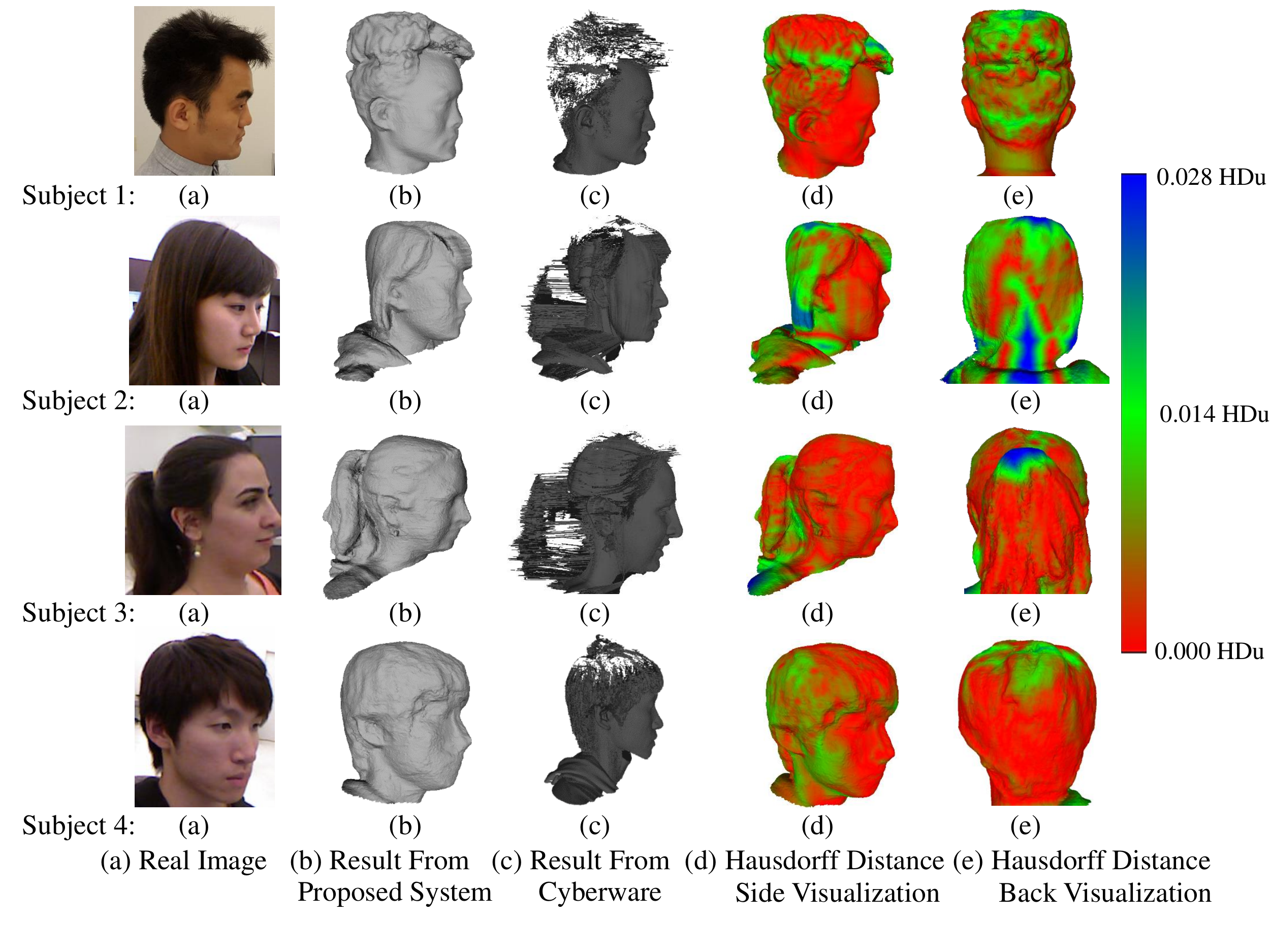}
	\caption{Male and female subjects' real image, scanned results from our proposed system and Cyberware separately, and the Hausdorff distance visualization results.}
	\label{fig:Real_all}
\end{figure*}

Detail comparison results are shown in Fig. \ref{fig:Real_all}. We demonstrated the real images, results from proposed system, results from Cyberware, and Hausdorff distance visualization results of two male subjects and two female subjects separately. As we can see from both Fig. \ref{fig:Real_evaluation} and \ref{fig:Real_all}, Cyberware can reconstruct certain regions with more details than the proposed system, in human eyes, eyebrows, ears, nose, and mouth, mainly because its closer capture distance, higher resolution, and more stable fixed rotation configuration. However, Cyberware laser scanner does not work well with optically uncooperative materials, such as human hairs. Thus, the back-view comparisons (Fig. \ref{fig:Real_all}(e)) show a larger Hausdorff distance (green and blue areas in the visualization images) than other views, especially for women with long hairs or ponytail. As a result, our proposed system has superior hair-style reconstruction capabilities compared with Cyberware system. Based on the facts that most facial comparison are in red color (small Hausdorff distance range), better hair-style reconstruction capabilities, lower cost, and more outstanding  mobility, the overall performance of our proposed system can be considered to be comparable with the expensive commercial Cyberware laser scanning system.


\section{Conclusion}
\label{sec:Conclusion}

In this paper, we proposed an automatic human head scanning-printing system, which provides a complete pipeline to scan, reconstruct, select, and print out 3D human heads. With our developed composite sensor, a tablet, and a robot, we realized movement freedom of the developed sensing device without human intervention. In addition, we proposed an automatic selection method for both standard head model and real human head. After a computational cost evaluation, and accuracy comparisons with two commercial 3D laser scanning systems, we showed that our system achieves comparable results with expensive commercial laser scanning systems, and could be an instrument in generating large-scale human shape databases for ergonomics, product design or anthropological studies at a much lower cost. 

For future work, we plan to utilize a stabilizer to increase the stability of Kinect v2 when capturing images under movement, and aim at porting our scanning-printing pipeline to the cloud, thus any individual that does not have the required processing power on their machine can simply send the raw data to a server and the final reconstructed human head can be digitally delivered in a matter of minutes.

\section*{Acknowledgments}
Longyu Zhang gratefully acknowledges the financial support from Natural Sciences and Engineering Research Council of Canada (NSERC) Postgraduate Doctoral Scholarship.
\bibliographystyle{unsrt}
\bibliography{literature}


\end{document}